\documentclass[letterpaper]{article} 
\usepackage[preprint]{aaai2027}
\usepackage[hyphens]{url}  
\usepackage{graphicx} 
\usepackage{natbib}  
\usepackage{caption} 
\usepackage{algorithm}
\usepackage{algorithmic}
\usepackage{xspace}
\usepackage{newfloat}
\usepackage{listings}
\usepackage{pifont}
\usepackage{makecell}
\usepackage{amsmath}
\usepackage{multirow}
\usepackage{enumitem}
\usepackage{soul}
\usepackage{colortbl}
\usepackage{fontawesome5}
\usepackage{placeins}
\usepackage{xspace}

\DeclareRobustCommand{\paperlink}[2]{%
  \leavevmode\hbox{%
    \pdfstartlink attr{/Border [0 0 0]} user{%
      /Subtype /Link
      /A << /S /URI /URI (\pdfescapestring{#1}) >>%
    }%
    #2%
    \pdfendlink
  }%
}
\newcommand{\huggingface}{\raisebox{-1.5pt}{\includegraphics[height=1.05em]{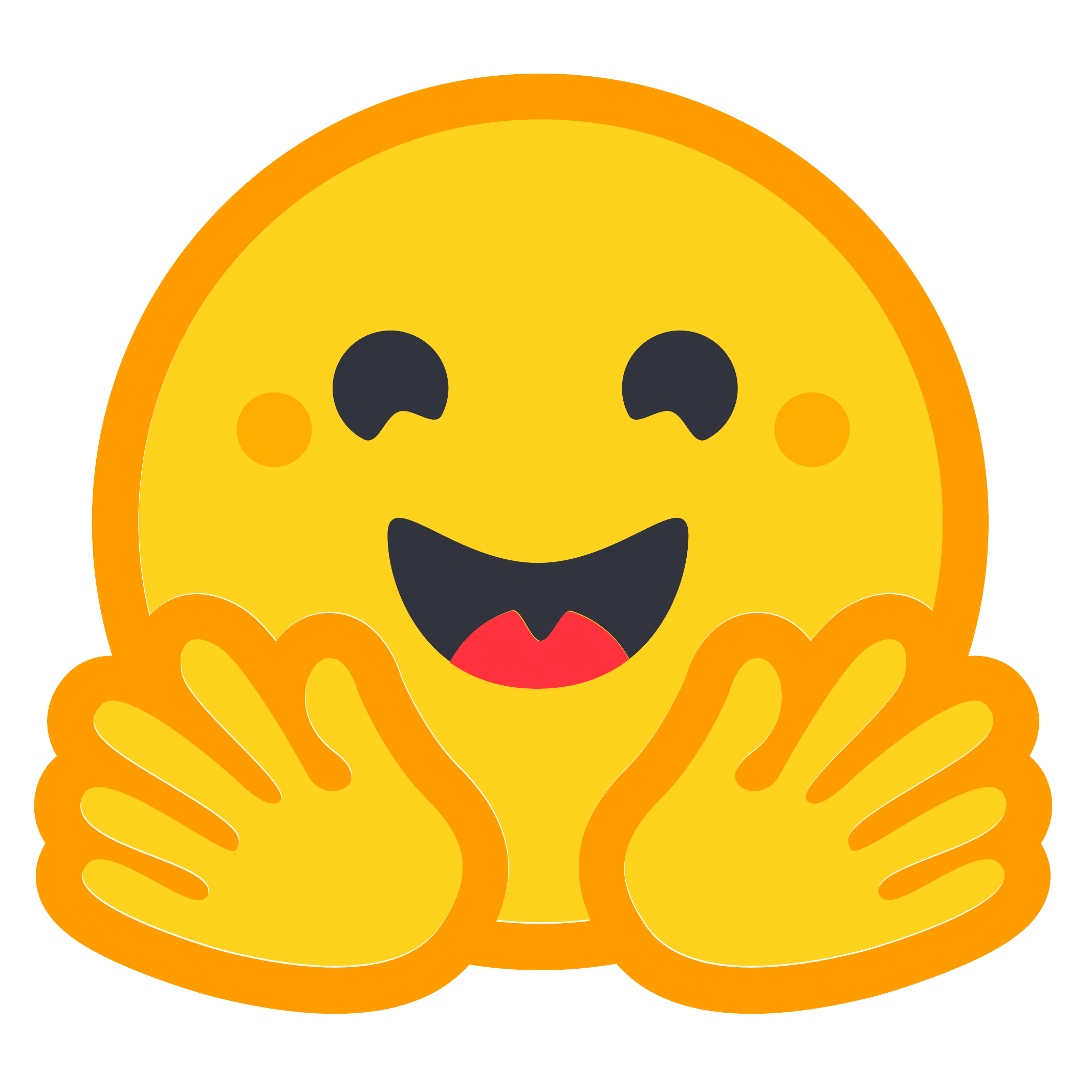}}\xspace}\newcommand{\github}{\raisebox{-1.5pt}{\includegraphics[height=1.05em]{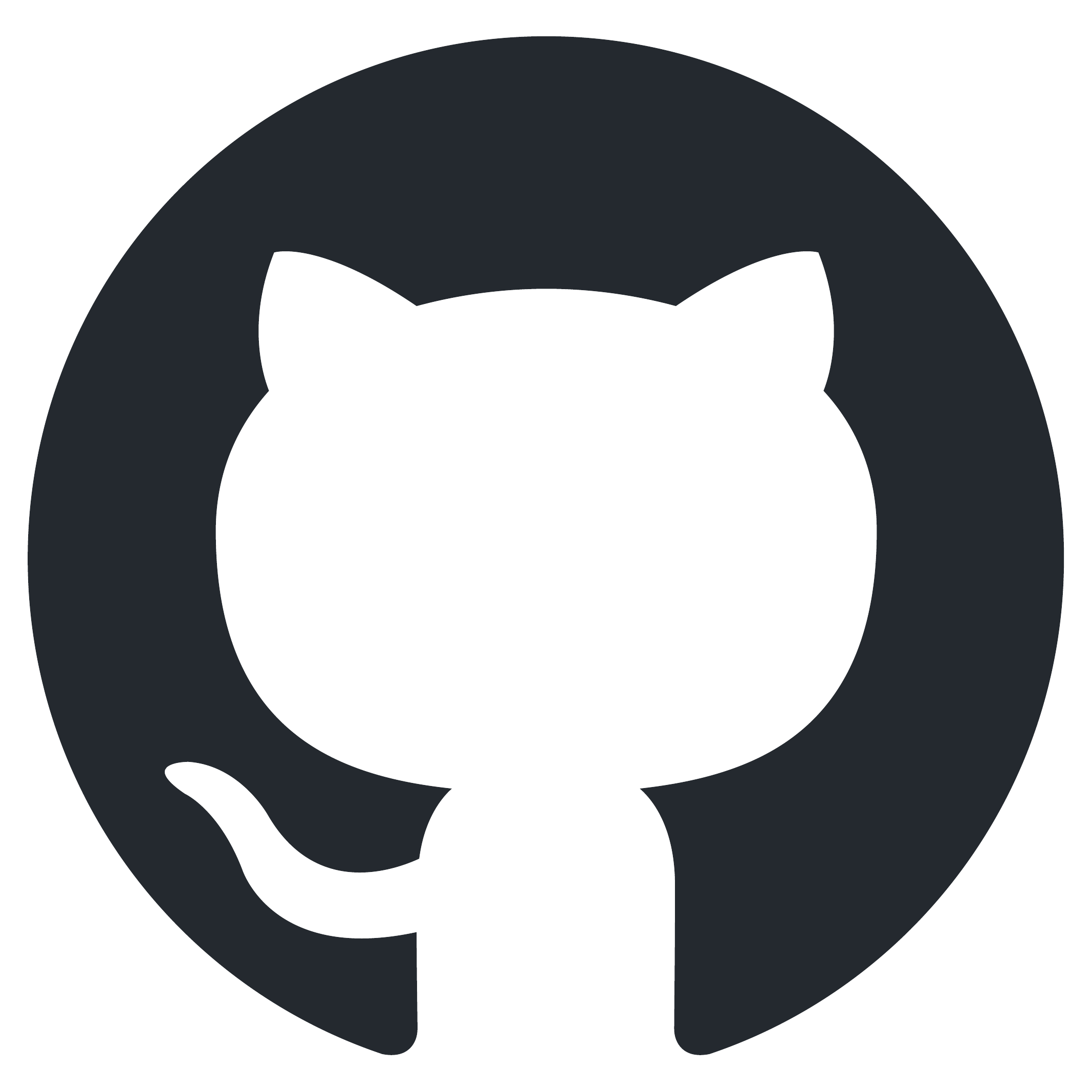}}\xspace}\newcommand{\modelscope}{\raisebox{0pt}{\includegraphics[height=0.8em]{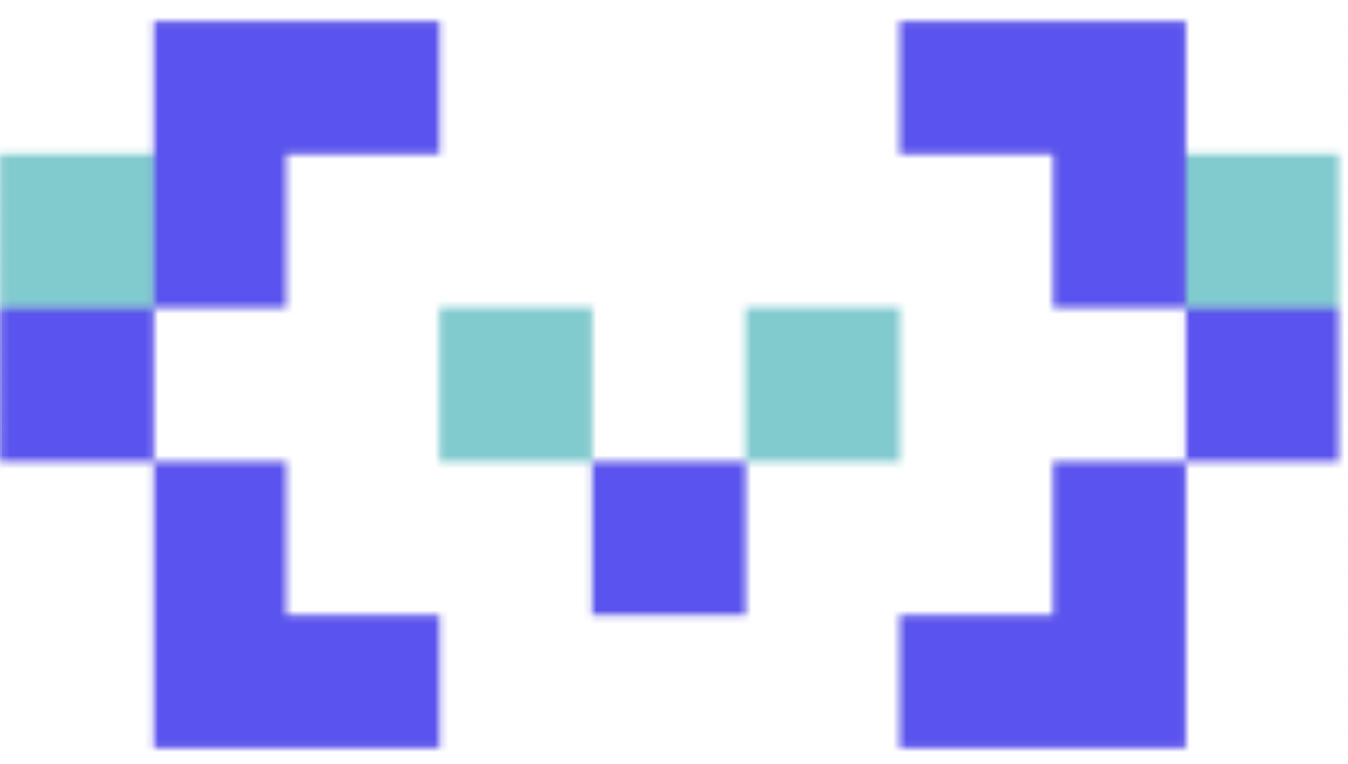}}\xspace}\newcommand{\project}{\raisebox{0pt}{\includegraphics[height=1.0em]{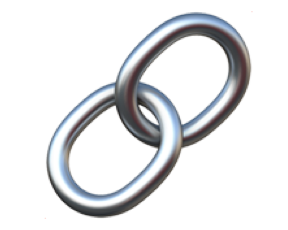}}\xspace}
\newcommand{\bench}{\textsc{R-GroundBench}}

\newcommand{\esmiles}{\textsc{E-SMILES}}

\newcommand{\cmark}{\ding{51}}
\newcommand{\xmark}{\ding{55}}

\newcommand{\first}[1]{\textbf{#1}}
\newcommand{\second}[1]{\underline{#1}}

\DeclareCaptionStyle{ruled}{labelfont=normalfont,labelsep=colon,strut=off} 
\floatstyle{ruled}
\newfloat{listing}{tb}{lst}{}
\floatname{listing}{Listing}

\usepackage{booktabs}

\title{\bench{}: A Diagnostic Benchmark for R-Group Grounding\\
in Markush Molecular Editing}

\author{
    Xin Wang\textsuperscript{\rm 1,*},
    Zichuan Ying\textsuperscript{\rm 1,2,*},
    Xinna Lin\textsuperscript{\rm 1,3,4},
    Junqi Zhang\textsuperscript{\rm 4},
    Hanyi Xiong\textsuperscript{\rm 1, 2},
    Tianyu Gao\textsuperscript{\rm 5},
    Hairong Zhang\textsuperscript{\rm 6},
    Qixiang Hua\textsuperscript{\rm 7},
    Botian Shi\textsuperscript{\rm 3,6},
    Zhenhailong Wang\textsuperscript{\rm 8},
    Kaicheng Yu\textsuperscript{\rm 1,3,†}
}

\affiliations{
    \textsuperscript{\rm 1}Westlake University\\
    \textsuperscript{\rm 2}The University of Hong Kong\\
    \textsuperscript{\rm 3}Shanghai Innovation Institute\\
    \textsuperscript{\rm 4}Zhejiang University\\
    \textsuperscript{\rm 5}Sichuan University\\
    \textsuperscript{\rm 6}Shanghai Artificial Intelligence Laboratory\\
    \textsuperscript{\rm 7}The Hong Kong University of Science and Technology (Guangzhou)\\
    \textsuperscript{\rm 8}University of Illinois Urbana-Champaign\\

    {\normalfont\normalsize
\paperlink
  {https://latent-recast-research.github.io/R-GroundBench}
  {\project{} Homepage}%
\hspace{1.2em}%
\paperlink
  {https://github.com/CrisYing/R-GROUNDBENCH/}
  {\github{} Code}%
\hspace{1.2em}%
\paperlink
  {https://huggingface.co/datasets/latent-recast-research/R-GroundBench}
  {\huggingface{} Hugging Face}%
\hspace{1.2em}%
\paperlink
  {https://www.modelscope.cn/datasets/Latent-Recast-Research/R-GroundBench}
  {\modelscope{} ModelScope}%
}
    \\
    \texttt{\{wangxin82, kyu\}@westlake.edu.cn,crisyingzc@gmail.com}
}

\begin{document}

\maketitle

\begin{abstract}
Recent advances in AI for scientific discovery enable molecular understanding
and design, yet reasoning over incomplete chemical representations remains
unclear.
Markush structures, which encode molecular families through variable R-group
placeholders (\textit{R\textsubscript{1}}, \textit{R\textsubscript{2}}, \textit{X}, etc.), are ubiquitous in pharmaceutical patents and require
grounding across molecular, textual, and chemical information.
However, existing molecule-language benchmarks focus on fully specified
molecules, leaving R-group grounding largely unevaluated.
We introduce \bench{}, a diagnostic benchmark built from real patent Markush
structures, featuring a Multiple-Choice (VQA) track with controlled difficulty
and modality splits, and an open-ended Generation track.
Our results reveal a substantial gap between recognition and
molecular grounding.
While models achieve over 90\% accuracy on Easy VQA, performance drops to
56--66\% on Hard VQA when shortcuts are controlled.
Chemical-domain VLMs also remain unreliable, achieving only 25.7--46.2\% on Hard
VQA despite domain-specific pretraining.
Moreover, Generation Exact Match remains below 20\% for most models and below
8\% when visual input is required.
These findings reveal that current AI systems lack reliable grounding and
execution for Markush editing, highlighting challenges for AI-driven
scientific discovery.

\end{abstract}

\section{Introduction}
\label{sec:intro}
\begin{figure*}[htb]
  \centering
   \vspace{-1.2em}
  \includegraphics[width=0.9\linewidth]{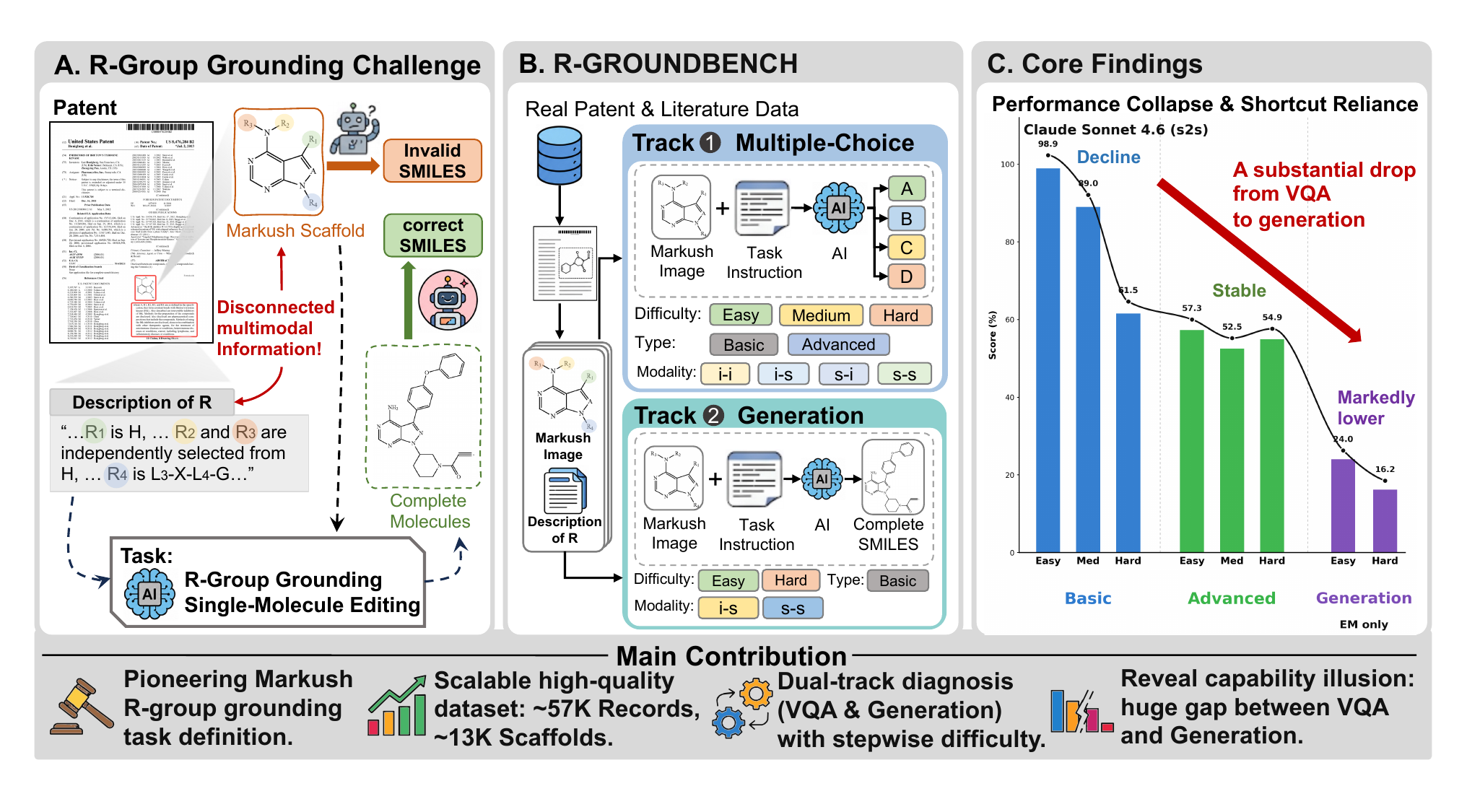}
   \vspace{-1.em}
  \caption{
Overview of \bench{}.
\textbf{A}: R-group grounding requires aligning disconnected molecular,
textual, and symbolic information.
\textbf{B}: \bench{} evaluates this capability through complementary VQA
and Generation tracks with controlled difficulty and modality settings.
\textbf{C}: Results reveal a gap between candidate recognition and
autonomous molecular instantiation, exposing shortcut reliance in current
models.
}
  \label{fig:overview}
  \vspace{-1.2em}
\end{figure*}

Foundation models have shown increasing potential for AI-assisted scientific
discovery, yet their ability to understand structured scientific objects and
perform precise domain-specific transformations remains unclear. Reliable
scientific AI requires more than generating plausible outputs: models must
interpret structured representations, connect information across different
modalities, follow symbolic design constraints, and faithfully execute
scientific transformations. These capabilities are essential for reliable
scientific AI, but remain largely unexplored in existing evaluations.

The rapid progress of molecular AI has produced benchmarks for optical chemical
structure recognition~\cite{molgrapher,molparser}, molecule captioning and
text-guided generation~\cite{molinstructions}, complete-molecule editing
~\cite{moleditrl}, and chemical reasoning~\cite{chemcotbench}. These benchmarks
have significantly advanced the evaluation of molecular understanding and
generation. However, they primarily focus on fully specified molecules, where
the complete molecular structure is already available. Such settings allow
models to operate on explicit molecular graphs without resolving latent
chemical variables. They therefore do not examine whether models can ground
abstract molecular placeholders to concrete chemical structures before
performing molecular instantiation. As a result, it remains unclear whether
current general-purpose language/vision-language models and chemical-domain
models possess genuine molecular grounding ability or rely on superficial
structural shortcuts.

We investigate this missing capability through Markush structures
~\cite{markush1924}, a representation widely used in pharmaceutical patents to
describe families of related molecules using variable placeholders
(e.g., \textit{R\textsubscript{1}}, \textit{R\textsubscript{2}}, and
\textit{X}). Unlike conventional molecular representations that specify a
single compound, Markush structures require models to associate abstract
substituent variables with their corresponding chemical realizations. We define
this capability as \emph{R-group grounding}: the ability to localize variable
positions, resolve intended substitutions from instructions, and instantiate
the corresponding molecular structures. Although fundamental to reliable
molecular editing, this capability has not been systematically evaluated by
existing benchmarks.

To address this gap, we introduce \bench{}, a diagnostic benchmark for
multimodal R-group grounded Markush molecular editing. An overview of the
benchmark motivation, evaluation tracks, and core findings is presented in
Figure~\ref{fig:overview}. \bench{} is constructed from real patent and
literature Markush structures sourced from MolParser-7M~\cite{molparser},
containing aligned molecular images, \esmiles{} representations,
natural-language editing instructions, edited structures, and molecular
annotations.

\bench{} evaluates models through two complementary settings. The
\textbf{Multiple-Choice (VQA)} track measures whether models can identify the
correct edited structure among candidates under controlled difficulty levels,
while the \textbf{Generation} track requires models to directly instantiate the
edited molecule without candidate options. Together, these settings distinguish
candidate-level recognition from end-to-end molecular editing ability. The
benchmark further supports four image/SMILES input-output modalities, enabling
systematic analysis of representation effects.

Our main contributions are:
\begin{itemize}[leftmargin=1.2em, nosep]
\item We introduce R-group grounded Markush molecular editing as a diagnostic
task for evaluating molecular grounding ability in foundation models,
connecting multimodal reasoning with AI for scientific discovery.

\item We construct \bench{} from the MolParser-7M \texttt{sft\_real} subset,
filtering 46{,}727 Markush-containing patent records into 14{,}394 clean
single-scaffold samples, resulting in 56{,}500 instruction records across
12{,}962 unique Markush scaffolds.

\item We benchmark general-purpose LLMs, VLMs, and chemical-domain models on
\bench{}, revealing a large gap between candidate recognition and autonomous
molecular instantiation: models perform strongly on simple VQA but degrade
substantially on hard grounding and generation.
\end{itemize}

\section{Related Work}
\label{sec:related}

\begin{table}[htb]
  \centering
  \scriptsize
  \setlength{\tabcolsep}{3pt}
   \vspace{-1.2em}
  \begin{tabular}{lccccc}
    \toprule
    \textbf{Benchmark}
    & \rotatebox{60}{\textbf{Markush}}
    & \rotatebox{60}{\textbf{R-grnd.}}
    & \rotatebox{60}{\textbf{Editing}}
    & \rotatebox{60}{\textbf{Multimod.}}
    & \rotatebox{60}{\textbf{Shortcut}} \\
    \midrule
    MolGrapher        & \xmark & \xmark & \xmark & img     & \xmark \\
    DECIMER           & \xmark & \xmark & \xmark & img     & \xmark \\
    MolParser         & \cmark & \xmark & \xmark & img     & \xmark \\
    MarkushGrapher    & \cmark & \xmark & \xmark & img+txt & \xmark \\
    MolLangBench      & \xmark & \xmark & \cmark & img+txt & \xmark \\
    Mol-Instructions  & \xmark & \xmark & \cmark & txt     & \xmark \\
    MolEditRL         & \xmark & \xmark & \cmark & txt     & \xmark \\
    ChemCoTBench      & \xmark & \xmark & \cmark & txt     & \xmark \\
    \midrule
    \bench{} (ours)    & \cmark & \cmark & \cmark & img+txt & \cmark \\
    \bottomrule
  \end{tabular}
   \vspace{-0.3em}
  \caption{%
    Comparison of related benchmarks.
    Markush: supports Markush structures;
    R-grnd.: evaluates R-group grounding;
    Editing: tests molecular editing;
    Multimod.: input modalities;
    Shortcut: diagnoses shortcut reasoning.
  }
  \label{tab:comparison}
  \vspace{-1.2em}
\end{table}

\begin{figure*}[htb]
  \centering
   \vspace{-1.2em}
  \includegraphics[width=0.9\linewidth]{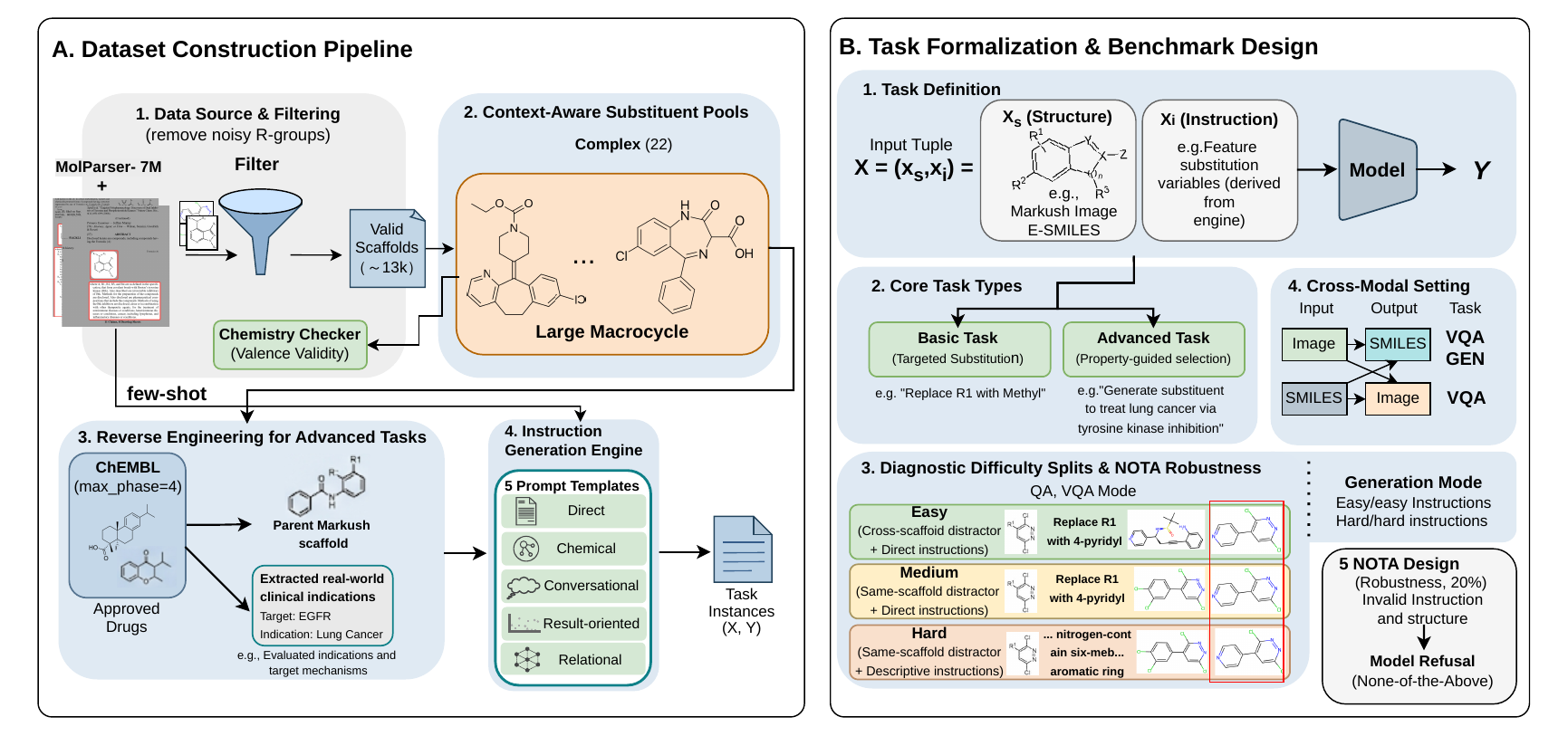}
   \vspace{-1.em}
   \caption{%
Overview of \bench{} construction and task formulation.
\textbf{A}: Dataset construction from real Markush structures with
filtering, annotation, and instruction generation.
\textbf{B}: Dual-track evaluation design, including VQA with controlled
difficulty/modality settings and Generation without candidate options.
}
  \label{fig:framework}
  \vspace{-1.2em}
\end{figure*}


\paragraph{Molecular Structure and Patent Analysis.}
Optical chemical structure recognition (OCSR) converts molecular images into
machine-readable representations such as SMILES~\cite{clide,osra}.
Recent methods, including MolGrapher~\cite{molgrapher}, DECIMER~\cite{decimer},
and MolParser~\cite{molparser}, improve molecular structure extraction from
images. In particular, MolParser introduces \esmiles{} to represent Markush
structures by preserving R-group placeholders alongside molecular scaffolds,
enabling machine-readable representations of patent-defined molecule families.
For pharmaceutical patent analysis, PatentAgent~\cite{patentagent} explores
agent-based document understanding, while MarkushGrapher~\cite{markushgrapher}
extracts backbone graphs and substituent tables from patent documents.
However, these approaches focus on molecular extraction or patent-level
analysis, whereas \bench{} evaluates the subsequent capability of grounding
R-group placeholders and performing instruction-guided molecular editing.

\paragraph{Molecule-Language Benchmarks.}
MolLangBench~\cite{mollangbench} covers language-prompted molecular recognition, editing, and generation. Mol-Instructions~\cite{molinstructions} provides data for captioning, generation, and property prediction.  MolEditRL~\cite{moleditrl} edit complete molecules via SMILES instructions. ChemCoTBench~\cite{chemcotbench} uses modular reasoning for molecular transformations. Unlike prior work on complete, fixed molecules, \bench{} targets variable Markush scaffolds, requiring grounding of placeholder symbols before instantiation, mirroring real patent workflows. Table~\ref{tab:comparison} summarizes the comparison between
R-GroundBench and existing molecular AI benchmarks.



\paragraph{Shortcut Reasoning in Multimodal Evaluation.}
VQA models may exploit language biases~\cite{vqacp} or visual
statistics~\cite{gvqa,pathvqa} instead of genuine understanding.
In molecular VQA, similar shortcuts can arise from matching substituent names
rather than grounding molecular structures. \bench{} mitigates this issue
through Easy/Medium/Hard splits, progressively removing such
shortcuts via scaffold-level and instruction-level constraints.
Beyond shortcut control, \bench{} addresses a limitation of complete-molecule
editing benchmarks~\cite{moleditrl}: Markush editing requires both placeholder
localization and substituent instantiation from natural-language instructions.
Errors in localization directly lead to incorrect molecular edits, making these
two capabilities essential for reliable R-group grounding.

\section{\bench{}}
\label{sec:benchmark}

\subsection{Data Construction}
\label{sec:benchmark:data}

R-GroundBench is constructed through a three-stage pipeline:
collecting and filtering real Markush structures, generating grounded
R-group editing instances, and enriching samples with molecular annotations
for advanced reasoning tasks.
The resulting benchmark provides aligned molecular images, E-SMILES
representations, editing instructions, edited structures, and property
annotations.
The overall benchmark construction pipeline and task formulation are
illustrated in Figure~\ref{fig:framework}.

\paragraph{\textbf{Step 1: Collecting and Filtering Markush Structures.}}
We start from the \texttt{sft\_real} subset of MolParser-7M~\citep{molparser},
containing 91{,}166 molecular image--E-SMILES pairs extracted from patent and
literature documents.
Validity filters remove ambiguous structures and invalid representations,
yielding 14{,}394 clean single-scaffold Markush structures.
Multi-scaffold structures are excluded due to ambiguity in R-group localization.

\paragraph{\textbf{Step 2: Constructing R-group Editing Instances.}}
For each valid Markush structure, we parse R-group positions from \esmiles{},
assign context-aware substituents, validate generated molecules with RDKit,
and render 300$\times$300 PNG images.
We further construct natural-language editing instructions from real patent
claim templates, ensuring that the benchmark reflects practical molecular
editing scenarios.
This process produces 55{,}982 unique editing operations across 12{,}962
scaffolds and 56{,}500 instruction records.
Detailed filtering criteria, substituent construction, and instruction
generation procedures are provided in Appendix.

\paragraph{\textbf{Step 3: Adding Molecular Annotations.}}
To support Advanced tasks, we enrich each editing instance with molecular
properties.
Six physicochemical descriptors are computed via RDKit (MW, logP, TPSA, HBD,
HBA, and rotatable bonds).
Biological annotations are obtained by reverse-engineering Markush structures
from approved drugs in ChEMBL~\cite{chembl}, linking molecules to bioactivity
pIC$_{50}$, indication, and mechanism of action records.
These annotations enable property-aware reasoning evaluation in the Advanced
task type. The key statistics of \bench{} are summarized in Appendix.

\subsection{Task Formulation}
\label{sec:benchmark:task}

\subsubsection{Definition and Two Tracks}
\label{sec:benchmark:formal}

Let $M$ denote a Markush representation, either a molecular image
$M_\text{img}$ or an \esmiles{} string $M_\text{smi}$.
\esmiles{} encodes a Markush structure as a backbone SMILES (with R-group
attachment points marked as \texttt{*}), followed by a \texttt{<sep>}
delimiter and XML-like tokens that map each attachment point to its
R-group label:
\begin{center}
  \footnotesize
  \texttt{N1=CC(*)=C(*)N=C1*<sep>}\\
  \texttt{<a>3:R[1]</a><a>5:R[2]</a><a>8:X</a>}
\end{center}
Here atoms 3, 5, and 8 in the backbone carry placeholders R[1], R[2], and X,
respectively.
$M_\text{img}$ and $M_\text{smi}$ represent two modalities of the same
Markush structure, and our benchmark evaluates models under both settings.

\bench{} evaluates R-group grounding through two complementary tracks that
probe different stages of molecular instantiation.
The Multiple-Choice (VQA) track evaluates candidate-level grounding by
providing four candidate molecules and asking the model to select
$C^*$ from $\mathcal{C}=\{C_1,\ldots,C_4\}$.
In contrast, the Generation track removes candidate options and requires
models to directly construct the edited molecule as a SMILES string given
$M$ and instruction $I$.
Together, the two tracks distinguish candidate recognition from
end-to-end molecular instantiation.

\subsubsection{Multiple-Choice (VQA) Track}
\label{sec:benchmark:vqa}


The VQA track evaluates candidate-level R-group grounding through two task
types and three difficulty levels.
Basic and Advanced tasks examine different reasoning requirements after
grounding, while Easy/Medium/Hard splits remove shortcut cues through
controlled distractors and instruction design.


\paragraph{Task types and difficulty splits.}
In the Basic task, $I$ specifies the desired substituent modification
(e.g., ``Replace R\textsubscript{1} with 4-chlorophenyl'').
The model must localize the placeholder in $M$, ground it to the intended
substituent, and select the candidate $C^*$ with the correct structural edit.
Basic tasks evaluate placeholder grounding and candidate instantiation without
property reasoning.

In the Advanced task, $I$ asks property-guided selection questions
(e.g., ``Which R-group is most hydrophobic?'').
Property questions cover three categories: Physicochemical (25\%, RDKit),
Biological activity (45\%, ChEMBL potency), and Indication/Mechanism (30\%,
ChEMBL approved drugs).
Advanced tasks evaluate property-aware selection after grounding rather than
property prediction.

We design three difficulty levels that progressively reduce shortcut cues by
controlling distractor and instruction style.
Easy uses cross-scaffold distractors with direct substituent names, allowing
lexical matching shortcuts; Medium uses same-scaffold distractors with direct
names to remove scaffold-level cues; Hard uses same-scaffold distractors with
chemically descriptive instructions, requiring models to infer the intended
substituent beyond lexical matching.
Importantly, Hard is not designed to measure general language complexity;
instead, it evaluates whether models can perform description-to-structure
grounding after removing associations between substituent names and
candidate structures.

\begin{table}[htb]
  \centering
  \scriptsize
  \vspace{-1.em}
  \begin{tabular}{llll}
    \toprule
    \textbf{Mode} & \textbf{Input} & \textbf{Candidates} & \textbf{Tests} \\
    \midrule
    i2i & Markush image & mol.\ images   & Visual grounding \\
    i2s & Markush image & SMILES options & Image$\to$symbol \\
    s2i & \esmiles{}    & mol.\ images   & Symbol$\to$visual \\
    s2s & \esmiles{}    & SMILES options & Symbolic editing \\
    \bottomrule
  \end{tabular}
  \vspace{-0.3em}
  \caption{The four input-output modality settings in \bench{}.
Notation: \textit{i2i} = image in, image candidates;
\textit{i2s} = image in, SMILES candidates;
\textit{s2i} = E-SMILES in, image candidates;
\textit{s2s} = E-SMILES in, SMILES candidates.
}
  \label{tab:modalities}
  \vspace{-1.2em}
\end{table}

\paragraph{Input-output modalities.}
To disentangle visual perception, symbolic manipulation, and cross-modal
alignment, the VQA track covers four input-output modality combinations that shown in
Table~\ref{tab:modalities}.
Text-only LLMs are evaluated solely under the s2s setting;
VLMs are evaluated across all four settings, enabling us to attribute
performance differences to visual perception failures, symbolic reasoning
failures, or cross-modal alignment failures.

\paragraph{NOTA robustness.}
To test candidate rejection, 20\% of Basic Easy/Medium questions include a
NOTA option: 5\% where NOTA is a distractor (correct among A/B/C);
10\% where all candidates contain structural errors (NOTA correct);
and 5\% where the instruction references a non-existent R-group (NOTA correct).
No NOTA questions appear in Hard splits, as Hard focuses on precise R-group
grounding among valid rather than invalid candidates.

\begin{table*}[!t]
  \centering
  \fontsize{9pt}{9pt}\selectfont
  \vspace{-1.2em}
  \setlength{\tabcolsep}{2.5pt}
  \begin{tabular}{@{}lllrrrrrrrrrrr@{}}
    \toprule
    \multirow{2}{*}{\textbf{Model Type}} & \multirow{2}{*}{\textbf{Model}} & \multirow{2}{*}{\textbf{Modality}}
    & \multicolumn{4}{c}{\textbf{Basic}}
    & \multicolumn{4}{c}{\textbf{Advanced}}
    & \multicolumn{2}{c}{\(\Delta_{\mathrm{EH}}\)}
    & \multirow{2}{*}{\textbf{Avg.}} \\
    \cmidrule(lr){4-7} \cmidrule(lr){8-11} \cmidrule(lr){12-13}
    & & 
    & Easy & Med & Hard & Avg.
    & Easy & Med & Hard & Avg.
    & Basic & Adv
    & \\
    \midrule


\multirow{32}{*}{General VLMs}

& \multirow{4}{*}{Qwen3-VL-8B~\cite{qwen3vl}}
& i2i & 25.5 & 24.9 & 25.1 & 25.2 & 21.3 & 20.4 & 21.1 & 20.9 & 0.4 & 0.2 & 23.0 \\
& & i2s & 93.6 & 74.8 & 55.7 & 74.7 & 55.2 & 54.7 & 55.7 & 55.2 & 37.9 & -0.5 & 64.9 \\
& & s2i & 78.4 & 68.7 & 49.6 & 65.6 & 37.7 & 38.2 & 37.2 & 37.7 & 28.8 & 0.5 & 51.6 \\
& & s2s & 93.4 & 72.7 & 55.9 & 74.0 & 49.4 & 48.7 & 49.2 & 49.1 & 37.5 & 0.2 & 61.6 \\

\cmidrule(l){2-14}

& \multirow{4}{*}{Qwen3-VL-32B~\cite{qwen3vl}}
& i2i & \second{98.1} & 90.2 & 55.5 & 81.3 & 44.6 & 44.2 & 47.0 & 45.3 & 42.6 & -2.4 & 63.3 \\
& & i2s & 97.7 & 91.3 & 60.9 & 83.3 & \second{60.4} & \second{60.3} & \first{61.0} & \second{60.6} & 36.8 & -0.6 & 71.9 \\
& & s2i & 95.2 & 90.6 & 55.6 & 80.5 & 46.4 & 47.2 & 44.0 & 45.9 & 39.6 & 2.4 & 63.2 \\
& & s2s & 97.5 & 91.4 & 61.9 & 83.6 & 50.3 & 48.9 & 49.1 & 49.4 & 35.6 & 1.2 & 66.5 \\

\cmidrule(l){2-14}

& \multirow{4}{*}{Claude Sonnet 4.6\cite{claude_sonnet_4_6}}
& i2i & \first{98.5} & \second{94.6} & 61.6 & \second{84.9} & 48.4 & 46.4 & 51.6 & 48.8 & 36.9 & -3.2 & 66.8 \\
& & i2s & \first{98.3} & \second{94.8} & 63.8 & 85.6 & 59.2 & 52.5 & 54.3 & 55.3 & 34.5 & 4.9 & 70.5 \\
& & s2i & \first{97.9} & 86.0 & 59.7 & 81.2 & 53.3 & 50.8 & 52.4 & 52.2 & 38.2 & 0.9 & 66.7 \\
& & s2s & \first{98.9} & 89.0 & 61.5 & 83.1 & \second{57.3} & \second{52.5} & 54.9 & \first{54.9} & 37.4 & 2.4 & 69.0 \\

\cmidrule(l){2-14}

& \multirow{4}{*}{GPT-4o~\cite{gpt4o}}
& i2i & 97.5 & 83.8 & 58.2 & 79.8 & 43.9 & 43.9 & 42.9 & 43.6 & 39.3 & 1.0 & 61.7 \\
& & i2s & 96.8 & 90.6 & 60.6 & 82.7 & 57.3 & 58.2 & 52.1 & 55.9 & 36.2 & 5.2 & 69.3 \\
& & s2i & \second{97.5} & 83.1 & 57.7 & 79.4 & 47.3 & 49.2 & 47.6 & 48.0 & 39.8 & -0.3 & 63.7 \\
& & s2s & 97.7 & 89.8 & 61.1 & 82.9 & 35.2 & 37.9 & 42.7 & 38.6 & 36.6 & -7.5 & 60.7 \\

\cmidrule(l){2-14}

& \multirow{4}{*}{GPT-4.1~\cite{gpt41}}
& i2i & 97.8 & 87.3 & 60.4 & 81.8 & 53.5 & 50.8 & 50.5 & 51.6 & 37.4 & 3.0 & 66.7 \\
& & i2s & 97.5 & 90.4 & 62.8 & 83.6 & 59.0 & \first{60.7} & \second{54.1} & 57.9 & 34.7 & 4.9 & 70.8 \\
& & s2i & 96.4 & 84.9 & 60.3 & 80.5 & 53.3 & 53.5 & 50.2 & 52.3 & 36.1 & 3.1 & 66.4 \\
& & s2s & 97.3 & 90.5 & 61.9 & 83.2 & 39.1 & 41.6 & 42.1 & 40.9 & 35.4 & -3.0 & 62.1 \\

\cmidrule(l){2-14}

& \multirow{4}{*}{GPT-5.5~\cite{gpt55}}
& i2i & 96.8 & \first{95.5} & \first{64.3} & \first{85.5} & \second{54.9} & \second{57.7} & \second{50.5} & \second{54.4} & 32.5 & 4.4 & \second{70.0} \\
& & i2s & \second{97.9} & \first{98.5} & \first{66.1} & \first{87.5} & 60.3 & 59.5 & 58.4 & 59.4 & 31.8 & 1.9 & \first{73.5} \\
& & s2i & 95.3 & \first{94.0} & \first{64.4} & \first{84.6} & \second{54.9} & \second{55.4} & \second{55.8} & \second{55.4} & 30.9 & -0.9 & \second{70.0} \\
& & s2s & \second{98.3} & \first{98.5} & \second{66.1} & \first{87.6} & \first{58.8} & 48.1 & \first{53.8} & 53.6 & 32.2 & 5.0 & \first{70.6} \\

\cmidrule(l){2-14}

& \multirow{4}{*}{Gemini 2.5 Flash~\cite{gemini25}}
& i2i & 95.3 & 93.8 & \second{62.6} & 83.9 & \first{59.8} & \first{58.2} & \first{62.2} & \first{60.1} & 32.7 & -2.4 & \first{72.0} \\
& & i2s & 97.5 & \second{94.8} & \second{65.4} & \second{85.9} & \first{62.5} & \second{60.3} & \second{60.1} & \first{61.0} & 32.1 & 2.4 & \second{73.4} \\
& & s2i & 96.8 & \second{93.4} & \second{62.0} & \second{84.1} & \first{61.4} & \first{58.7} & \first{61.4} & \first{60.5} & 34.8 & 0.0 & \first{72.3} \\
& & s2s & 98.1 & \second{95.1} & 63.8 & \second{85.7} & 54.7 & \first{53.5} & \second{52.5} & \second{53.6} & 34.3 & 2.2 & \second{69.6} \\

\cmidrule(l){2-14}

& \multirow{4}{*}{Llama 4 Maverick~\cite{llama4}}
& i2i & 92.7 & 86.2 & 59.5 & 79.5 & 35.2 & 36.0 & 33.3 & 34.8 & 33.2 & 1.9 & 57.2 \\
& & i2s & 93.6 & 82.0 & 62.6 & 79.4 & 56.3 & 54.4 & 55.8 & 55.5 & 31.0 & 0.5 & 67.5 \\
& & s2i & 89.7 & 81.2 & 59.9 & 76.9 & 36.0 & 38.2 & 37.5 & 37.2 & 29.8 & -1.5 & 57.1 \\
& & s2s & 90.6 & 82.3 & 63.7 & 78.9 & 33.6 & 36.3 & 37.7 & 35.9 & 26.9 & -4.1 & 57.4 \\

\midrule


\multirow{5}{*}{General LLMs}

& DeepSeek-V4-Pro~\cite{deepseekv4}
& s2s 
& 97.3 & \first{98.1} & \second{68.7} & \second{88.0}
& \first{52.5} & \first{55.1} & \first{54.6} & \first{54.1}
& 28.6 & 13.4 & \first{71.0} \\

& DeepSeek-R1~\cite{deepseekr1}
& s2s 
& 94.0 & 95.0 & 62.6 & 83.9
& 47.6 & \second{54.7} & 42.0 & 48.1
& 31.4 & 5.6 & 66.0 \\

& LLaMA-3.3-70B-Instruct~\cite{llama3}
& s2s 
& 96.7 & 79.6 & 60.4 & 78.9
& 42.6 & 42.9 & 44.2 & 43.2
& 36.3 & -1.6 & 61.1 \\

& o3~\cite{o3}
& s2s 
& \second{97.7} & \second{97.7} & \first{69.3} & \first{88.2}
& 44.5 & 43.1 & \second{52.5} & 46.7
& 28.4 & -8.0 & 67.5 \\

& Qwen3-Max-2025-09-23~\cite{qwen3}
& s2s 
& \first{98.6} & 95.5 & 64.2 & 86.1
& \second{47.9} & 49.2 & 50.9 & \second{49.3}
& 34.4 & -3.0 & \second{67.7} \\

\midrule


\multirow{8}{*}{\makecell{Chemical-domain\\Models}}

& \multirow{4}{*}{ChemVLM-8B~\cite{chemvlm}}
& i2i 
& 34.2 & 36.0 & 28.9 & 33.0
& 23.7 & 21.8 & 22.9 & 22.8
& 11.3 & 0.8 & 27.9 \\

& & i2s 
& 96.8 & 85.9 & 56.7 & 79.8
& 63.0 & 63.4 & 63.8 & 63.4
& 33.0 & -0.4 & 71.6 \\

& & s2i 
& 62.2 & 53.9 & 40.2 & 52.1
& 35.8 & 38.5 & 37.4 & 37.2
& 24.8 & -0.5 & 44.7 \\

& & s2s 
& 96.3 & 84.2 & 58.8 & 79.7
& 61.2 & 60.4 & 61.5 & 61.0
& 34.8 & 0.7 & 70.4 \\

\cmidrule(l){2-14}

& \multirow{4}{*}{MolVL-7B~\cite{molvl}}
& i2i 
& 22.0 & 20.5 & 25.5 & 22.7
& 22.1 & 29.4 & 24.0 & 25.2
& -2.0 & -3.9 & 23.9 \\

& & i2s 
& 20.5 & 19.6 & 25.7 & 21.9
& 22.0 & 29.7 & 23.9 & 25.2
& -3.4 & -4.0 & 23.6 \\

& & s2i 
& 21.7 & 19.5 & 25.2 & 22.1
& 21.9 & 30.0 & 23.8 & 25.2
& -2.1 & -3.9 & 23.7 \\

& & s2s 
& 22.5 & 22.8 & 25.7 & 23.7
& 21.9 & 29.8 & 23.8 & 25.2
& -1.3 & -4.0 & 24.4 \\

\bottomrule

  \end{tabular}
\vspace{-0.3em}
  \caption{%
  Full VQA results on \bench{} (\% accuracy). Columns: Basic/Advanced accuracy by difficulty (Easy/Med/Hard), their means, \(\Delta_{\mathrm{EH}}\) (Easy minus Hard), and overall mean across six splits. Rows: each model $\times$ modality (i2i, i2s, s2i, s2s); LLMs on s2s only. \first{bold}: best; \second{underline}: second best among models within each modality and model-type block. Chemical-domain models are not marked because model group contains fewer than three models.
  }
  \label{tab:main}
\end{table*}

\subsubsection{Generation Track}
\label{sec:benchmark:gen}

The Generation track evaluates end-to-end molecular instantiation by
requiring models to directly generate edited SMILES without candidate
options.
Unlike VQA, which measures candidate selection under controlled settings,
Generation evaluates whether models can autonomously ground R-groups,
realize molecular edits, and produce valid molecular representations.

\paragraph{Task design.}
The task requires models to complete the full editing pipeline, including
R-group grounding, molecular realization, and SMILES generation.
Although SMILES is the output representation, evaluation focuses on whether
models preserve the Markush scaffold and perform instruction-grounded
R-group substitution correctly.

We restrict Generation to Basic tasks because they directly evaluate molecular
instantiation after substituent grounding.
In contrast, Advanced tasks further involve property-aware reasoning, and
introducing Generation in this setting would entangle failures in property
understanding with failures in molecular realization.
Generation therefore serves as a complementary evaluation setting following
the grounding progression established in VQA.

Difficulty is controlled by instruction style:
Easy uses direct substituent names, while Hard uses chemically descriptive
instructions that require inference beyond explicit names.
We evaluate two input modalities: s2s and i2s.

\section{Experiments and Results}
\label{sec:exp}

\subsection{Experimental Setup}
\label{sec:exp:setup}

\begin{table}[htb]
  \centering
  \fontsize{9pt}{8pt}\selectfont
  \setlength{\tabcolsep}{4.0pt}
  \begin{tabular}{@{}l ccc ccc@{}}
    \toprule
    & \multicolumn{3}{c}{\textbf{Easy}} & \multicolumn{3}{c}{\textbf{Hard}} \\
    \cmidrule(lr){2-4} \cmidrule(lr){5-7}
    \textbf{Model}
    & EM & Scaff. & Tani. & EM & Scaff. & Tani. \\
    \midrule

    \multicolumn{7}{l}{\textit{VLMs --- s2s}}\\
    \addlinespace[2pt]

    Qwen3-VL-8B
    & 4.2 & 49.2 & 44.0
    & 3.0 & 52.2 & 45.6 \\

    Qwen3-VL-32B
    & 12.8 & 38.0 & 57.7
    & 10.0 & 41.4 & 52.9 \\

    Claude Sonnet 4.6
    & \second{24.0} & 31.2 & \first{88.6}
    & \second{16.2} & 23.0 & \first{87.5} \\

    GPT-4o
    & 19.4 & 53.0 & 59.9
    & 10.4 & 50.8 & 54.0 \\

    GPT-4.1
    & 15.4 & 48.2 & 57.3
    & 9.8 & 51.0 & 53.1 \\

    GPT-5.5
    & \first{44.6} & \first{62.2} & \second{78.4}
    & \first{38.4} & \first{64.0} & \second{76.7} \\

    Gemini 2.5 Flash
    & 20.4 & \second{55.8} & 62.5
    & 20.8 & \second{56.4} & 62.6 \\

    Llama 4 Maverick
    & 12.2 & 48.6 & 52.0
    & 9.0 & 51.2 & 48.9 \\

    \addlinespace[4pt]

    \multicolumn{7}{l}{\textit{VLMs --- i2s}}\\
    \addlinespace[2pt]

    Qwen3-VL-8B
    & 1.2 & 25.0 & 22.3
    & 0.1 & 25.6 & 23.3 \\

    Qwen3-VL-32B
    & 2.6 & 31.0 & 34.1
    & 2.6 & 33.0 & 32.6 \\

    Claude Sonnet 4.6
    & \second{10.2} & 23.8 & \first{66.5}
    & \second{7.8} & 17.4 & \first{72.9} \\

    GPT-4o
    & 4.4 & 31.4 & 36.5
    & 1.8 & 23.2 & 34.4 \\

    GPT-4.1
    & 4.2 & 31.4 & 33.8
    & 1.8 & 27.6 & 35.4 \\

    GPT-5.5
    & \first{16.2} & \first{51.8} & \second{59.8}
    & \first{17.0} & \first{54.0} & \second{61.3} \\

    Gemini 2.5 Flash
    & 8.2 & \second{42.2} & 51.1
    & 5.6 & \second{40.4} & 48.2 \\

    Llama 4 Maverick
    & 7.0 & 45.0 & 43.1
    & 5.2 & 44.4 & 40.8 \\

    \midrule

    \multicolumn{7}{l}{\textit{LLMs --- s2s}}\\
    \addlinespace[2pt]

    DeepSeek-V4-Pro
    & \first{50.7} & \first{72.7} & \first{82.6}
    & \first{29.7} & \second{54.6} & \first{71.7} \\

    DeepSeek-R1
    & 24.5 & 45.5 & \second{73.1}
    & 20.3 & 45.9 & 64.0 \\

    LLaMA-3.3-70B
    & 4.8 & 35.6 & 47.1
    & 1.6 & 38.2 & 39.5 \\

    o3
    & \second{37.1} & \second{64.7} & 76.6
    & \second{28.7} & \first{64.1} & \second{72.2} \\

    Qwen3-Max
    & 15.6 & 50.0 & 58.1
    & 13.8 & 48.6 & 56.2 \\

    \midrule

    \multicolumn{7}{l}{\textit{Chemical-domain Models --- i2s}}\\
    \addlinespace[2pt]

    MarkushGrapher-2
    & 5.4 & 14.8 & 68.6
    & 0.0 & 0.0 & 0.0 \\

    ChemVLM-8B
    & 0.6 & 37.8 & 32.8
    & 0.6 & 41.6 & 30.9 \\

    MolVL-7B
    & 0.0 & 28.4 & 18.0
    & 0.0 & 24.6 & 19.1 \\

    \addlinespace[4pt]

    \multicolumn{7}{l}{\textit{Chemical-domain Models --- s2s}}\\
    \addlinespace[2pt]

    ChemVLM-8B
    & 2.8 & 34.0 & 41.4
    & 1.8 & 36.2 & 36.9 \\

    MolVL-7B
    & 0.0 & 18.2 & 19.5
    & 0.0 & 15.2 & 16.8 \\

    \bottomrule
  \end{tabular}
  \caption{%
    Generation track results (\%).
    EM = Exact Match, Scaff.\ = Scaffold Match, Tani.\ = Tanimoto Similarity.
    Chemical-domain models are not marked.
  }
  \label{tab:gen}
\end{table}

\textbf{Models.}
We evaluate three categories of models: general-purpose Large Language
Models, general-purpose vision-language models, and chemical-domain models specialized in molecular
understanding or structure processing.
The latter includes chemical VLMs (ChemVLM-8B and MolVL-7B) and the
Markush-specific model MarkushGrapher-2.
Model details and citations are listed in Table~\ref{tab:main}.

\paragraph{\textbf{Evaluation Metrics.}}
\label{sec:benchmark:metrics}

For the VQA track, we report split-wise accuracy for Basic tasks and
property-type accuracy for Advanced tasks.
We define $\Delta_{\text{EH}} =
\text{Acc}_{\text{Easy}}-\text{Acc}_{\text{Hard}}$ to quantify the
performance drop after shortcut removal, and report NOTA accuracy on
Easy/Medium Basic splits.
For the Generation track, we report three metrics with increasing tolerance
to generation errors.
Exact Match (EM) is the primary metric after canonicalizing predictions and
ground truth.
Scaffold Match measures backbone preservation using Maximum Common
Substructure (MCS), requiring $\geq$80\% scaffold heavy-atom coverage.
Tanimoto Similarity provides a continuous structural similarity measure when
EM fails.

\paragraph{\textbf{Baselines.}}
For VQA, random selection among four candidates provides a 25\%
chance-level baseline.
For Generation, we introduce a rule-based \textit{template oracle} as a
non-learning reference to quantify performance achievable through
explicit substituent-name matching.
The oracle extracts substituent names from Easy instructions, retrieves
corresponding SMILES from a lookup table, and performs RDKit-based
substitution without learned representation or reasoning.
The oracle achieves 88.0\% EM and 0.969 Tanimoto on Easy Generation,
confirming solvability.
For Hard Generation, chemically descriptive instructions contain no explicit
substituent names that can be directly matched (parse rate~=~0\%), resulting
in 0.0\% EM for the oracle.
Although selected molecular editing cases are verified by chemistry experts
during benchmark construction, we do not provide a large-scale human
performance baseline due to the difficulty of recruiting sufficient
qualified evaluators with expertise in Markush structures and molecular
editing.
This contrast demonstrates that Hard Generation removes name-matching
shortcuts and requires instruction-grounded molecular instantiation rather
than explicit substituent retrieval.

\subsection{VQA Results and Analysis}

Table~\ref{tab:main} summarizes VQA performance across models, modalities,
and difficulty levels. Three findings emerge:

\begin{itemize}[leftmargin=*]

\item \textbf{Finding 1: Easy accuracy is high but shortcut-sensitive.}
Most VLMs exceed 90\% accuracy on Easy-Basic, but performance drops to
50--66\% on Hard-Basic with same-scaffold distractors and
shortcut-resistant instructions. For example, average s2i performance across
eight VLMs decreases from 93.4\% to 58.6\%, suggesting that Easy results
mainly reflect candidate discrimination rather than R-group grounding.
Under s2s Hard-Basic, LLMs remain competitive with VLMs (64.1\% vs.\
61.4\%), with o3 achieving the highest score (69.3\%), showing that reasoning
can partially compensate for missing visual input.

\item \textbf{Finding 2: Representation alignment dominates modality effects.}
Across four input-output modalities, Hard-Basic performance follows:
i2s (62.2) $\approx$ s2s (62.0) $>$ i2i (60.3) $>$ s2i (58.6).
The comparable performance between i2s and s2s suggests that visual input
itself is not the primary bottleneck. Instead, the main challenge lies in
aligning molecular structures, symbolic representations, and editing
operations, particularly when models must translate between symbolic and
visual representations.


\item \textbf{Finding 3: Chemical-domain models do not guarantee
R-group grounding ability.}
Chemical-domain VLMs provide a
test of whether molecular pretraining transfers to R-group grounding.
Although they are trained on chemical representations, they achieve
substantially lower performance than general-purpose VLMs on Basic VQA.
This indicates that molecular representation learning alone does not
guarantee reliable grounding of substituent instructions to candidate
structures.
\end{itemize}

\begin{figure*}[htb]
  \centering
    \vspace{-1.2em}
  \includegraphics[width=0.9\linewidth]{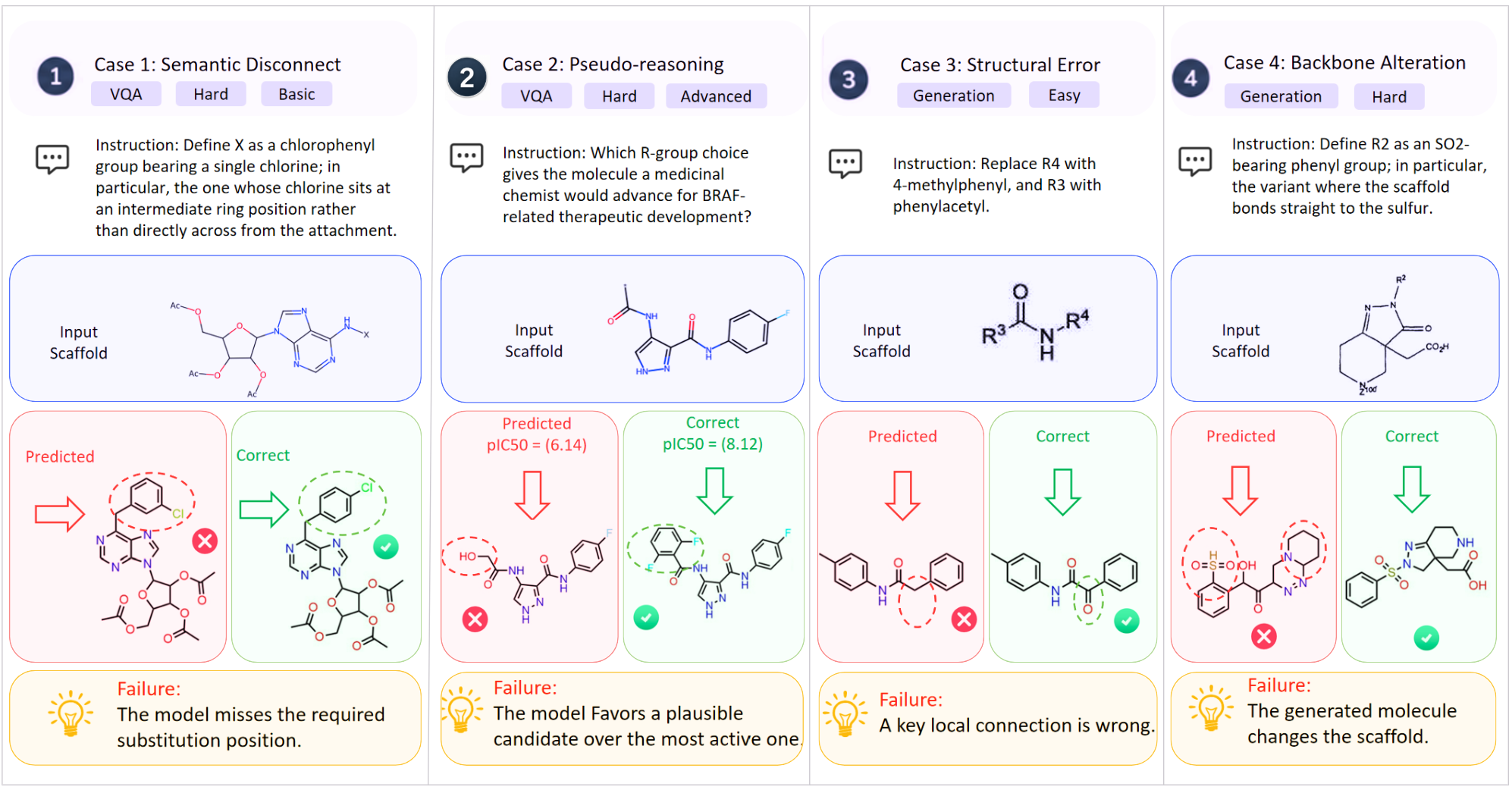}
  \vspace{-0.5em}
  \caption{Four representative failure cases of GPT-4.1 on \bench{}.}
  \label{fig:case_study}
  \vspace{-1.2em}
\end{figure*}


\paragraph{NOTA Robustness.}
We evaluate candidate rejection using NOTA questions (details in
Appendix). 
Results show that invalid candidate rejection
remains challenging. For example, GPT-4.1 achieves 98.8\% on NOTA-as-distractor
but only 76.2\% when structural errors require selecting NOTA, suggesting that
reliable validation requires explicit structural checking rather than
answer-selection heuristics.


\paragraph{Property-aware Chemical Reasoning.}
We analyze Advanced VQA by property type. Bioactivity questions are the most
challenging, with VLMs and LLMs achieving only 26--28\%, close to the 25\%
random baseline, likely due to the difficulty of inferring experimental
activity without assay-specific or SAR context. Physicochemical and indication
questions are easier due to computable properties and memorized knowledge,
showing that property-aware reasoning remains challenging beyond basic
R-group substitution.

\subsection{Generation Results and Analysis}
\label{sec:exp:gen}

Table~\ref{tab:gen} reveals a substantial gap between VQA selection and
instruction-conditioned molecular generation.
While VQA provides explicit candidates that reduce the search space,
Generation requires autonomous R-group grounding, molecular instantiation,
and SMILES construction. Although previous molecular instruction-following
studies have shown that LLMs can generate valid molecular representations
~\cite{molinstructions}, \bench{} Generation additionally requires resolving
which R-group to modify and how to instantiate the corresponding substituent.
The average VLM Exact Match decreases from 59.7\% in Hard-Basic VQA to
14.7\% (s2s) and 5.2\% (i2s) in Hard Generation.

\paragraph{Recognition does not translate into molecular instantiation.}
The large VQA-to-Generation gap indicates that candidate-level recognition
does not directly translate into executable editing ability.
When candidates are provided, models can often identify plausible
structures; however, they struggle to independently resolve R-group
assignments and construct the final molecular representation.

\paragraph{Symbolic input remains easier than visual generation.}
Across VLMs, s2s consistently outperforms image-based generation:
average EM decreases from 19.1\% to 6.8\% on Easy and from 14.7\% to
5.2\% on Hard when moving from s2s to i2s.
This suggests that visual input introduces additional alignment challenges,
while the core difficulty remains precise structural modification.


\paragraph{Chemical-domain models do not guarantee
instruction-grounded R-group generation.}
MarkushGrapher-2~\cite{markushgrapher2}, despite being specifically designed for Markush 
structure reconstruction, only addresses part of the required pipeline and does not
perform instruction-conditioned R-group substitution. It therefore achieves
limited performance when required to instantiate R-group instructions into
complete molecules. These results suggest that task-specific molecular
generation capability does not necessarily imply robust instruction-grounded
molecular editing.


\subsection{Failure Case Studies}
Four representative failure cases are shown in Figure~\ref{fig:case_study}.




\textbf{Case 1: Semantic Disconnect.}
The model identifies the correct chlorophenyl category but selects the
wrong substitution position, confusing an intermediate-position chlorine
with a para substitution.
This reveals a gap between linguistic interpretation and fine-grained
R-group grounding.

\textbf{Case 2: Pseudo-reasoning.}
In a BRAF-related bioactivity task, the model selects candidate A
(pIC$_{50}$=6.14) instead of candidate C (pIC$_{50}$=8.52), despite a
plausible rationale.
This indicates that fluent explanations do not necessarily reflect
reliable property-aware comparison.

\textbf{Case 3: Structural Error.}
The model generates a valid SMILES with moderate similarity to the target,
but introduces incorrect local connectivity and carbonyl patterns.
The generated molecule appears chemically plausible and preserves part of the
scaffold, yet fails to realize the requested R-group substitution.
This demonstrates the gap between producing plausible molecules and
performing precise molecular instantiation.

\textbf{Case 4: Backbone Alteration.}
The model modifies the molecular backbone instead of applying the requested
R-group substitution, resulting in a structurally different molecule.
This highlights the challenge of preserving the Markush scaffold during
autonomous editing.

\section{Conclusion}
\label{sec:conclusion}

We introduced \bench{}, a diagnostic benchmark for evaluating R-group
grounding in Markush molecular editing.
Our results reveal that current molecular AI capabilities are often
overestimated when evaluation only requires candidate recognition.
Although models achieve strong Easy VQA performance, they degrade under
harder grounding conditions and instruction-conditioned Generation, exposing a gap between
recognizing plausible structures and autonomously instantiating valid
molecules.

Beyond a dataset, \bench{} provides a diagnostic framework for separating
recognition, grounding, and execution through complementary evaluations,
including the $\Delta_\text{EH}$ gap and VQA$\rightarrow$Generation drop.
These analyses show that current molecular AI systems still rely on
superficial cues and lack structure-aware grounding for
molecular editing.

Our findings further identify two major challenges for future molecular AI
systems. Visual information alone does not guarantee reliable molecular
instantiation, while Advanced bioactivity reasoning remains difficult due
to the need to integrate structural understanding with empirical chemical
knowledge.
Addressing these challenges requires models that can jointly ground
visual, textual, and symbolic chemical information and execute chemically
precise transformations.
Such grounded scientific foundation models will be essential for reliable
AI-assisted molecular discovery and design.





\bibliography{custom}

@inproceedings{molparser,
  title={MolParser: End-to-end visual recognition of molecule structures in the wild},
  author={Fang, Xi and Wang, Jiankun and Cai, Xiaochen and Chen, Shangqian and Yang, Shuwen and Tao, Haoyi and Wang, Nan and Yao, Lin and Zhang, Linfeng and Ke, Guolin},
  booktitle={Proceedings of the IEEE/CVF International Conference on Computer Vision},
  pages={24528--24538},
  year={2025}
}

@inproceedings{molgrapher,
  title={MolGrapher: graph-based visual recognition of chemical structures},
  author={Morin, Lucas and Danelljan, Martin and Agea, Maria Isabel and Nassar, Ahmed and Weber, Valery and Meijer, Ingmar and Staar, Peter and Yu, Fisher},
  booktitle={Proceedings of the IEEE/CVF International Conference on Computer Vision},
  pages={19552--19561},
  year={2023}
}

@article{decimer,
  title     = {{DECIMER} 2.0: Deep Learning for Chemical Image Recognition using Transformers},
  author    = {Rajan, Kohulan and Zielesny, Achim and Steinbeck, Christoph},
  journal   = {Journal of Cheminformatics},
  volume    = {15},
  number    = {1},
  pages     = {37},
  year      = {2023}
}

@article{osra,
  title     = {{OSRA}: An Optical Structure Recognition Application},
  author    = {Filippov, Igor V. and Nicklaus, Marc C.},
  journal   = {Journal of Chemical Information and Modeling},
  volume    = {49},
  number    = {3},
  pages     = {740--743},
  year      = {2009}
}

@article{clide,
  title={CLiDE Pro: the latest generation of CLiDE, a tool for optical chemical structure recognition},
  author={Valko, Aniko T and Johnson, A Peter},
  journal={Journal of chemical information and modeling},
  volume={49},
  number={4},
  pages={780--787},
  year={2009},
  publisher={ACS Publications}
}

@inproceedings{markushgrapher,
  title     = {{MarkushGrapher}: Joint Visual and Textual Recognition of {Markush} Structures},
  author    = {Morin, Lucas and Weber, Valery and Nassar, Ahmed and Meijer, Gerhard Ingmar and Van Gool, Luc and Li, Yawei and Staar, Peter},
  booktitle = {Proceedings of the IEEE/CVF Conference on Computer Vision and Pattern Recognition (CVPR)},
  year      = {2025}
}

@misc{markush1924,
  title   = {Process of dyeing},
  author  = {Markush, Eugene A.},
  year    = {1924},
  note    = {{US} Patent 1,506,316. The first Markush-type patent claim}
}

@article{moleditrl,
  title={MolEditRL: Structure-Preserving Molecular Editing via Discrete Diffusion and Reinforcement Learning},
  author={Zhuang, Yuanxin and Shen, Dazhong and Sun, Ying},
  journal={arXiv preprint arXiv:2505.20131},
  year={2025}
}

@article{molinstructions,
  title     = {{Mol-Instructions}: A Large-Scale Biomolecular Instruction Dataset for Large Language Models},
  author    = {Fang, Yin and Liang, Xiaozhuan and Zhang, Ningyu and Liu, Kangwei and Huang, Rui and Chen, Zhuo and Fan, Xiaohui and Chen, Huajun},
  journal   = {arXiv preprint arXiv:2306.08018},
  year      = {2023}
}

@article{chemcotbench,
  title={Beyond chemical qa: Evaluating llm's chemical reasoning with modular chemical operations},
  author={Hao, Li and Cao, He and Feng, Bin and Shao, Daniel and Tang, Robert and Yan, Zhiyuan and Tian, Yonghong and Yuan, Li and Li, Yu},
  journal={Advances in Neural Information Processing Systems},
  volume={38},
  year={2026}
}

@misc{mollangbench,
      title={MolLangBench: A Comprehensive Benchmark for Language-Prompted Molecular Structure Recognition, Editing, and Generation}, 
      author={Feiyang Cai and Jiahui Bai and Tao Tang and Guijuan He and Joshua Luo and Tianyu Zhu and Srikanth Pilla and Gang Li and Ling Liu and Feng Luo},
      year={2026},
      eprint={2505.15054},
      archivePrefix={arXiv},
      primaryClass={cs.CL},
      url={https://arxiv.org/abs/2505.15054}, 
}

@misc{patentagent,
      title={$\texttt{PatentAgent}$: Intelligent Agent for Automated Pharmaceutical Patent Analysis}, 
      author={Xin Wang and Yifan Zhang and Xiaojing Zhang and Longhui Yu and Xinna Lin and Jindong Jiang and Bin Ma and Kaicheng Yu},
      year={2024},
      eprint={2410.21312},
      archivePrefix={arXiv},
      primaryClass={cs.LG},
      url={https://arxiv.org/abs/2410.21312}, 
}

@inproceedings{vqacp,
  title     = {Making the {V} in {VQA} Matter: Elevating the Role of Image Understanding in Visual Question Answering},
  author    = {Goyal, Yash and Khot, Tejas and Summers-Stay, Douglas and Batra, Dhruv and Parikh, Devi},
  booktitle = {Proceedings of the IEEE/CVF Conference on Computer Vision and Pattern Recognition (CVPR)},
  year      = {2017}
}

@inproceedings{gvqa,
  title     = {Don't Just Assume; Look and Answer: Overcoming Priors for Visual Question Answering},
  author    = {Agrawal, Aishwarya and Batra, Dhruv and Parikh, Devi and Kembhavi, Aniruddha},
  booktitle = {Proceedings of the IEEE/CVF Conference on Computer Vision and Pattern Recognition (CVPR)},
  year      = {2018}
}

@article{pathvqa,
  title     = {{PathVQA}: 30000+ Questions for Medical Visual Question Answering},
  author    = {He, Xuehai and Zhang, Yichen and Mou, Luntian and Xing, Eric and Xie, Pengtao},
  journal   = {arXiv preprint arXiv:2003.10286},
  year      = {2020}
}

@article{chembl,
  title={ChEMBL: a large-scale bioactivity database for drug discovery},
  author={Gaulton, Anna and Bellis, Louisa J and Bento, A Patricia and Chambers, Jon and Davies, Mark and Hersey, Anne and Light, Yvonne and McGlinchey, Shaun and Michalovich, David and Al-Lazikani, Bissan and others},
  journal={Nucleic acids research},
  volume={40},
  number={D1},
  pages={D1100--D1107},
  year={2012},
  publisher={Oxford University Press}
}

@article{deepseekv4,
  title     = {{DeepSeek-V4}: Towards Highly Efficient Million-Token Context Intelligence},
  author    = {{DeepSeek-AI}},
  journal   = {arXiv preprint arXiv:2606.19348},
  year      = {2026}
}

@article{deepseekr1,
  title     = {{DeepSeek-R1}: Incentivizing Reasoning Capability in {LLMs} via Reinforcement Learning},
  author    = {{DeepSeek-AI}},
  journal   = {arXiv preprint arXiv:2501.12948},
  year      = {2025}
}

@article{llama3,
  title     = {The {Llama 3} Herd of Models},
  author    = {{Meta AI}},
  journal   = {arXiv preprint arXiv:2407.21783},
  year      = {2024}
}

@misc{llama4,
  title        = {{Llama 4}: The Next Generation of {Meta}'s Open Models},
  author       = {{Meta AI}},
  year         = {2025},
  howpublished = {\url{https://ai.meta.com/blog/llama-4-multimodal-intelligence/}}
}

@article{qwen3vl,
  title     = {{Qwen3-VL} Technical Report},
  author    = {{Qwen Team}},
  journal   = {arXiv preprint arXiv:2511.21631},
  year      = {2025}
}

@article{qwen3,
  title     = {{Qwen3} Technical Report},
  author    = {{Qwen Team}},
  journal   = {arXiv preprint arXiv:2505.09388},
  year      = {2025}
}

@misc{gpt4o,
  title        = {{GPT-4o} System Card},
  author       = {{OpenAI}},
  year         = {2024},
  howpublished = {\url{https://openai.com/index/gpt-4o-system-card/}}
}

@misc{gpt41,
  title        = {{GPT-4.1}},
  author       = {{OpenAI}},
  year         = {2025},
  howpublished = {\url{https://openai.com/index/gpt-4-1/}}
}

@misc{gpt55,
  title        = {{GPT-5.5}},
  author       = {{OpenAI}},
  year         = {2025},
  howpublished = {\url{https://openai.com/}}
}

@misc{o3,
  title        = {{o3} System Card},
  author       = {{OpenAI}},
  year         = {2025},
  howpublished = {\url{https://openai.com/index/o3-system-card/}}
}

@misc{gemini25,
  title        = {{Gemini 2.5 Flash}},
  author       = {{Google DeepMind}},
  year         = {2025},
  howpublished = {\url{https://deepmind.google/models/gemini/flash/}}
}

@misc{claude_sonnet_4_6,
  title        = {{Claude Sonnet 4.6}},
  author       = {{Anthropic}},
  year         = {2026},
  howpublished = {\url{https://www.anthropic.com/news/claude-sonnet-4-6}}
}

@misc{chemvlm,
      title={ChemVLM: Exploring the Power of Multimodal Large Language Models in Chemistry Area}, 
      author={Junxian Li and Di Zhang and Xunzhi Wang and Zeying Hao and Jingdi Lei and Qian Tan and Cai Zhou and Wei Liu and Yaotian Yang and Xinrui Xiong and Weiyun Wang and Zhe Chen and Wenhai Wang and Wei Li and Shufei Zhang and Mao Su and Wanli Ouyang and Yuqiang Li and Dongzhan Zhou},
      year={2025},
      eprint={2408.07246},
      archivePrefix={arXiv},
      primaryClass={cs.LG},
      url={https://arxiv.org/abs/2408.07246}, 
}

@misc{molvl,
      title={OCSU: Optical Chemical Structure Understanding for Molecule-centric Scientific Discovery}, 
      author={Siqi Fan and Yuguang Xie and Bowen Cai and Ailin Xie and Gaochao Liu and Mu Qiao and Jie Xing and Zaiqing Nie},
      year={2025},
      eprint={2501.15415},
      archivePrefix={arXiv},
      primaryClass={cs.CV},
      url={https://arxiv.org/abs/2501.15415}, 
}

@misc{markushgrapher2,
      title={MarkushGrapher-2: End-to-end Multimodal Recognition of Chemical Structures}, 
      author={Tim Strohmeyer and Lucas Morin and Gerhard Ingmar Meijer and Valéry Weber and Ahmed Nassar and Peter Staar},
      year={2026},
      eprint={2603.28550},
      archivePrefix={arXiv},
      primaryClass={cs.CV},
      url={https://arxiv.org/abs/2603.28550}, 
}

\appendix












\section{Markush Filtering Criteria}
\label{app:filter}

We filter the 91{,}166 samples from MolParser-7M \texttt{sft\_real} according
to four criteria, applied sequentially. Table~\ref{tab:filter} reports the
number of samples remaining after each step.

\begin{enumerate}
  \item \textbf{Presence of R-group tokens}: the \esmiles{} must contain at
    least one standard placeholder (\texttt{R}, \texttt{R[1]}--\texttt{R[9]},
    \texttt{X}, \texttt{Y}, \texttt{Z}).
  \item \textbf{Clean token set}: samples containing non-standard tokens such
    as \texttt{<dum>}, \texttt{<r>}, or \texttt{<c>} are excluded.
  \item \textbf{R-group count}: samples with more than 10 R-group positions
    are excluded for complexity control.
  \item \textbf{RDKit validity}: the backbone SMILES (with R-groups replaced
    by wildcard atoms) must be parseable by RDKit.
\end{enumerate}

\begin{table}[t]
\centering
\small
\begin{tabular}{lrr}
\toprule
\textbf{Filter step} & \textbf{Remaining} & \textbf{Removed} \\
\midrule
Raw MolParser-7M \texttt{sft\_real}         & 91{,}166 & --- \\
\quad $+$ Has \texttt{<a>} tag (pre-filter) & 46{,}727 & 44{,}439 \\
\quad $+$ Standard R-group token only       & 19{,}282 & 27{,}445 \\
\quad $+$ No non-standard tokens            & 14{,}476 & 4{,}806  \\
\quad $+$ R-group count $\leq 10$           & 14{,}394 &      82  \\
\quad $+$ RDKit-parseable backbone          & 14{,}394 &       0  \\
\bottomrule
\end{tabular}
\caption{Sequential filtering statistics. The pre-filter (row 2) is applied
during dataset download and retains only samples containing at least one
\texttt{<a>} annotation tag; the remaining four steps correspond to the
criteria listed above.}
\label{tab:filter}
\end{table}

This yields 14{,}394 clean samples as the core source, from which
task construction generates 55{,}982 unique editing operations across
12{,}962 Markush scaffolds. Multi-scaffold structures --- where the backbone
SMILES contains disconnected components---are not used, as they introduce
ambiguity in VQA construction: it is undefined which scaffold component
the instruction refers to.

\section{Substituent Pool Design}
\label{app:pool}

All substituents are drawn from 22 structurally complex, ring-containing groups
(e.g., 4-chlorophenyl, 2-pyridyl, phenylsulfonyl), organised into seven similarity
clusters (Table~\ref{tab:subst_groups}). To prevent chemically invalid products,
we define three context-aware sub-pools: a \textbf{halogen-position pool} (7 halogenated
aryl/benzyl groups) for X/Y/Z placeholders, an \textbf{N-substituent pool} (16 groups)
for nitrogen attachment points, and an \textbf{O-substituent pool} (10 groups) for
oxygen attachment points---each a strict subset of the general pool.

\begin{table*}[t]
\centering
\small
\begin{tabular}{c c l}
\toprule
\textbf{Group} & \textbf{Size} & \textbf{Members} \\
\midrule
G0 & 5 & phenyl, 4-F-phenyl, 4-Cl-phenyl, 4-Me-phenyl, 4-OMe-phenyl \\
G1 & 3 & 3-chlorophenyl, 3,4-dichlorophenyl, 4-(trifluoromethyl)phenyl \\
G2 & 3 & 2-pyridyl, 3-pyridyl, 4-pyridyl \\
G3 & 3 & cyclohexyl, cyclopentyl, cyclopropyl \\
G4 & 3 & benzyl, 4-fluorobenzyl, 4-chlorobenzyl \\
G5 & 3 & benzoyl, phenylacetyl, 4-fluorobenzoyl \\
G6 & 2 & phenylsulfonyl, 4-methylsulfonylphenyl \\
\bottomrule
\end{tabular}
\caption{Seven similarity clusters of the 22-substituent pool.
Within-group members share a common scaffold but differ in ring size,
substitution pattern, or heteroatom position, providing challenging
fine-grained alternatives for Hard splits.}
\label{tab:subst_groups}
\end{table*}

\section{Instruction Styles}
\label{app:instructions}

Table~\ref{tab:instruct_styles} shows the five instruction templates
used in \bench{}. Three direct-name templates (Direct, Conversational,
and Result-oriented) are used for Easy and Medium splits, as they provide
explicit substituent names.
Two chemically grounded templates (Relational and Descriptive) are
reserved for Hard splits. Instead of directly specifying the substituent
name, these templates require models to infer the intended R-group from
structural relations or chemical descriptions.

\begin{table}[t]
  \centering
  \small
  \setlength{\tabcolsep}{3pt}
  \begin{tabular}{llp{3.2cm}}
    \toprule
    \textbf{Style} & \textbf{Split} & \textbf{Example} \\
    \midrule
    Direct          & Easy/Med & Replace R\textsubscript{1} with 4-methylphenyl. \\
    Conversational  & Easy/Med & Can you change R\textsubscript{1} to 4-methylphenyl? \\
    Result-oriented & Easy/Med & I need the R\textsubscript{1} position to be 4-methylphenyl. \\
    \midrule
    Relational      & Hard & Among the four candidates, select the one whose R\textsubscript{1} substituent contains the greatest number of halogen atoms. \\
    Chemically descriptive & Hard & Replace R\textsubscript{1} with a para-halogenated phenyl ring, specifically the one whose halogen is fluorine rather than chlorine. \\
    \bottomrule
  \end{tabular}
  \caption{Five instruction templates used in \bench{}. The upper three
           are direct-name styles deployed in Easy and Medium splits.
           The lower two are Hard templates that require additional
           grounding beyond explicit substituent names:
           \textbf{Relational} identifies the target through relations
           among candidates, while \textbf{Chemically descriptive}
           specifies the target through structural descriptions rather
           than direct naming.}
  \label{tab:instruct_styles}
\end{table}

The two Hard templates target complementary forms of R-group grounding.
The \textbf{Relational} template does not provide an explicit substituent
name or an isolated description of the target group; instead, it identifies
the correct candidate through comparisons among the four R-group fragments
(e.g., the greatest number of halogen atoms or the only nitrogen-containing
ring). Such instructions require models to reason over the candidate set
rather than rely on direct name matching, and they are generated only when
the specified relation is uniquely satisfied by one candidate.

The \textbf{Chemically descriptive} template replaces explicit substituent
names with chemically meaningful descriptions. It may introduce shared
properties among multiple candidates (e.g., ``a para-halogenated phenyl
ring'') and provide additional structural constraints to identify the target
(e.g., ``the one whose halogen is fluorine rather than chlorine''). Therefore,
successful prediction requires mapping chemical descriptions to the
corresponding molecular structure instead of matching surface-level names.

For scaffolds with multiple R-positions, one clause is generated per
position and the clauses are concatenated (``At R\textsubscript{1},
\ldots; and at R\textsubscript{2}, \ldots''). For every Hard question we
verify that exactly one of the four candidates satisfies all constraints;
questions failing this uniqueness check are discarded.


\begin{figure*}[thb]
  \centering
  \includegraphics[width=\linewidth]{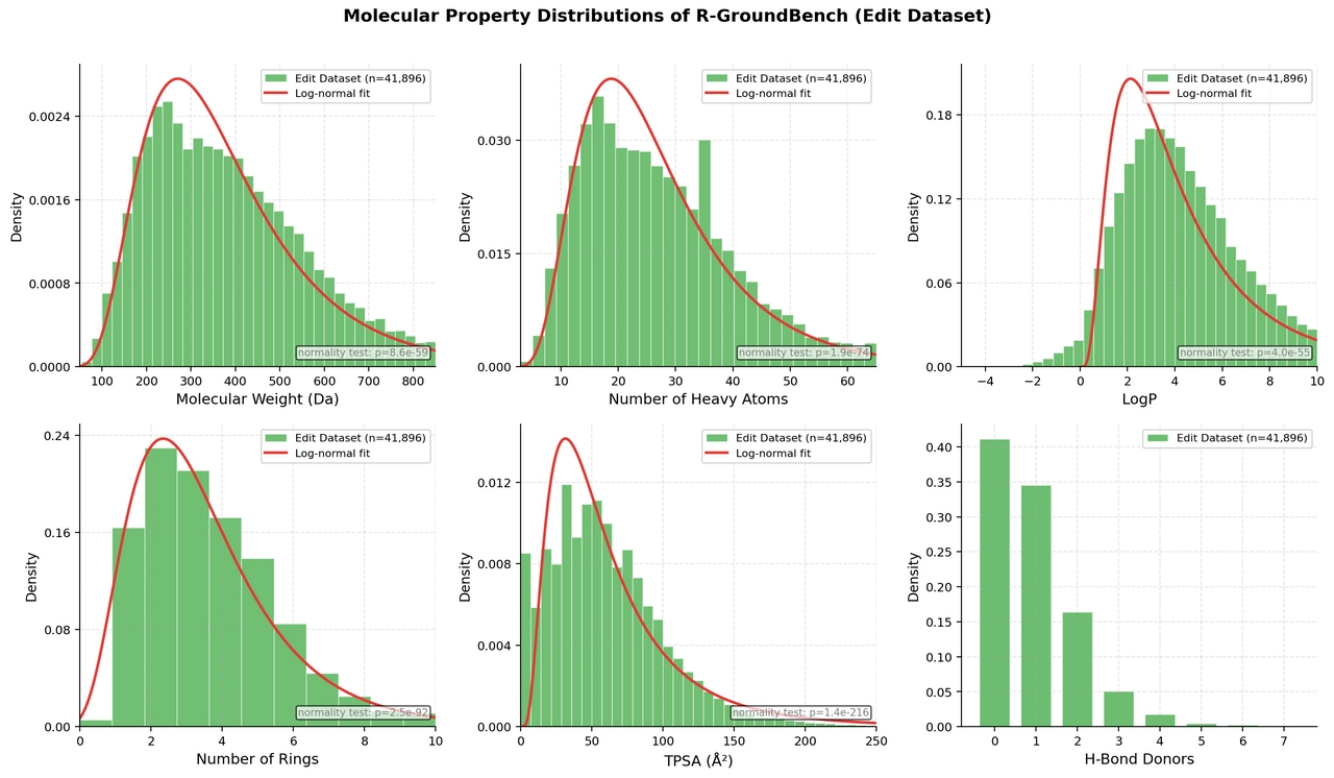}
  \caption{Physicochemical property distributions of edited molecules
    ($n = 55{,}982$). Green histograms with log-normal fits (red).}
  \label{fig:properties}
\end{figure*}

\section{Dataset Statistics and Diversity Analysis}
\label{app:stats}

\begin{table}[t]
  \centering
  \small
  \setlength{\tabcolsep}{4pt}
  \begin{tabular}{@{}lp{4.0cm}@{}}
    \toprule
    \textbf{Statistic} & \textbf{Value} \\
    \midrule
    Source subset              & MolParser-7M \texttt{sft\_real} \\
    Raw patent/lit.\ mols      & 91{,}166 \\
    Markush-containing         & 46{,}727 \\
    Clean samples used         & 14{,}394 \\
    Unique Markush scaffolds   & 12{,}962 \\
    Unique editing operations  & 55{,}982 \\
    Total instruction records  & 56{,}500 \\
    Basic VQA questions        & 940 per split \\
    Advanced VQA questions     & 634 per split \\
    NOTA questions             & ${\approx}$20\% of Easy/Med Basic \\
    Generation questions       & 500 easy + 500 hard \\
    Input modalities           & 4 (i2i / i2s / s2i / s2s) \\
    ChEMBL SAR clusters        & 308 \\
    Bioactivity data points    & 2{,}387 (mean pIC$_{50}$\,=\,6.92) \\
    Biological target classes  & 13 \\
    \bottomrule
  \end{tabular}
  \caption{\bench{} dataset statistics.}
  \label{tab:stats}
\end{table}

\paragraph{Physicochemical property distributions.}
Figure~\ref{fig:properties} shows that the 55{,}982 edited molecules
exhibit drug-like profiles: MW mean 428\,Da with 70\% in 300--700\,Da;
heavy atom count, LogP, ring count (1--3 dominant), TPSA ($<$100\,\AA$^2$),
and H-bond donors all follow log-normal distributions consistent with
oral drug space.

\begin{figure*}[t]
  \centering
  \includegraphics[width=\linewidth]{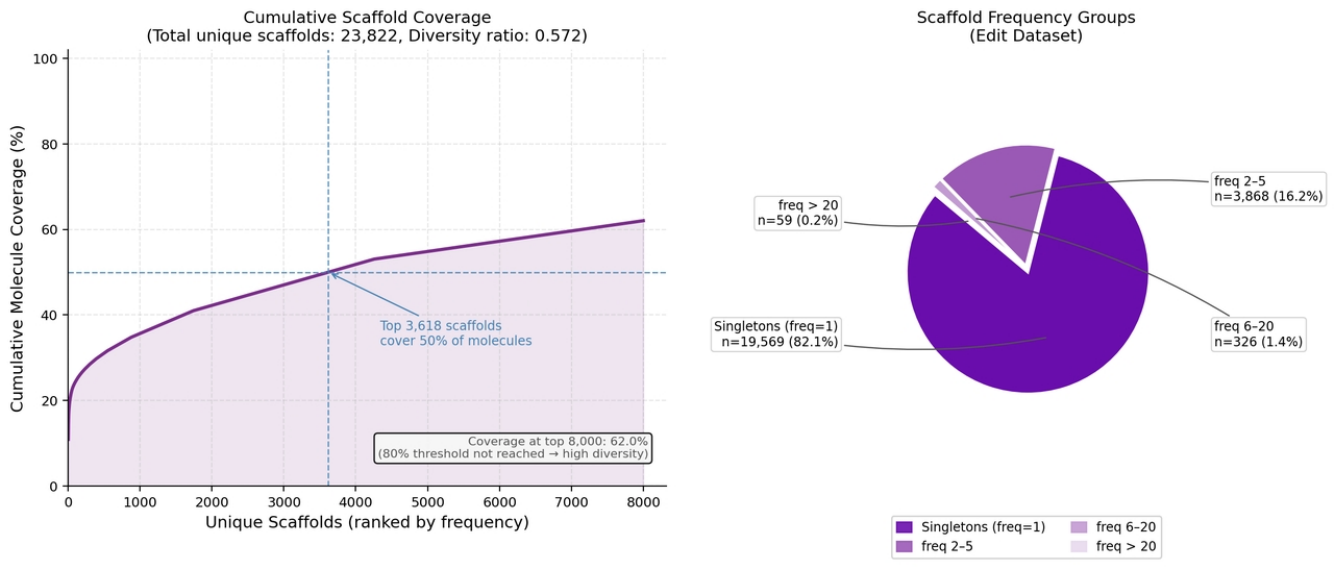}
  \caption{Scaffold diversity. \textbf{Left}: cumulative coverage curve
    (23{,}822 unique scaffolds). \textbf{Right}: frequency group
    distribution (82.1\% singletons).}
  \label{fig:scaffold}
\end{figure*}

\paragraph{Scaffold diversity.}
Murcko scaffold analysis (Figure~\ref{fig:scaffold}) reveals 23{,}822
unique scaffolds.
Singletons account for 82.1\% ($n = 19{,}569$); the top 3{,}618
scaffolds cover only 50\% of molecules, and even the top 8{,}000 reach
only 62\%---confirming no chemotype bias.

\begin{figure*}[t]
  \centering
  \includegraphics[width=\linewidth]{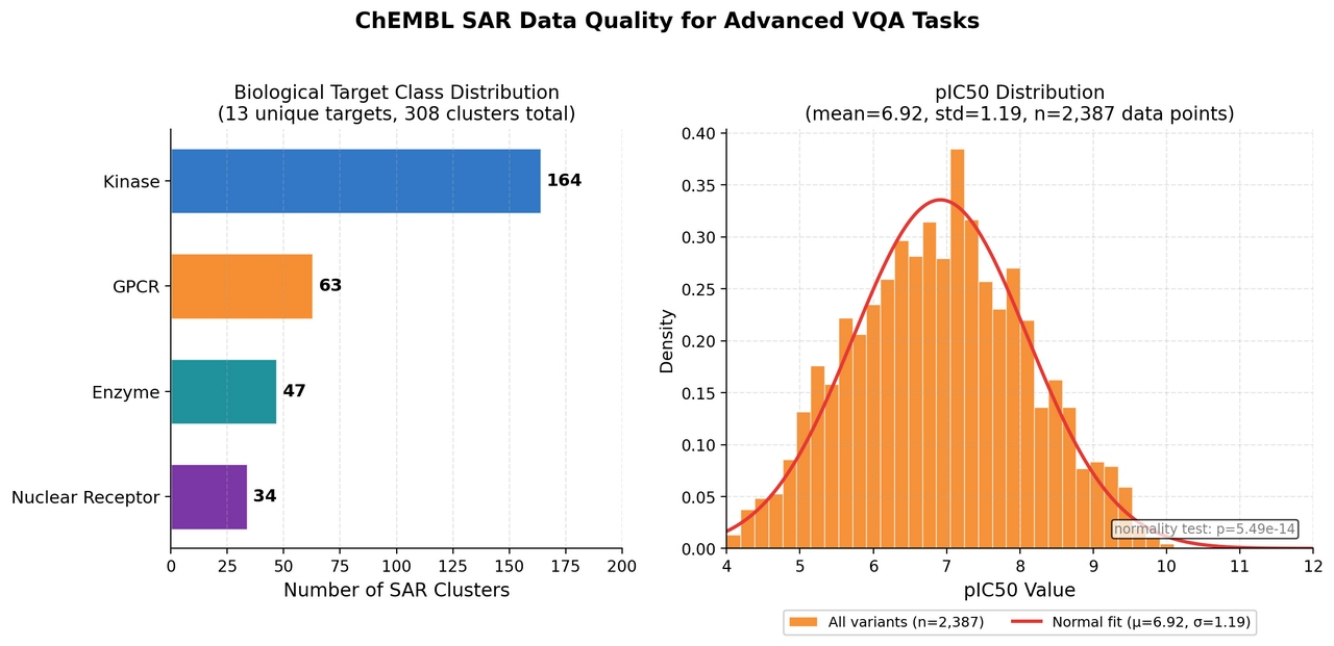}
  \caption{ChEMBL SAR data quality. \textbf{Left}: target class
    distribution (308 clusters, 13 types). \textbf{Right}: pIC$_{50}$
    distribution ($\mu = 6.92$, $\sigma = 1.19$, $n = 2{,}387$).}
  \label{fig:chembl}
\end{figure*}

\paragraph{ChEMBL Advanced data quality.}
The 308 SAR clusters span 13 target classes (Kinase 164, GPCR 63,
Enzyme 47, Nuclear Receptor 34; Figure~\ref{fig:chembl}).
The 2{,}387 pIC$_{50}$ values follow a near-normal distribution
($\mu = 6.92$, $\sigma = 1.19$), covering the full micromolar-to-sub-nanomolar
potency range relevant to drug discovery.

\section{From VQA to Generation: A Cross-Track Performance Cliff}
\label{app:gen_cliff}
\begin{figure*}[!t]
  \centering
  \includegraphics[width=0.90\linewidth]{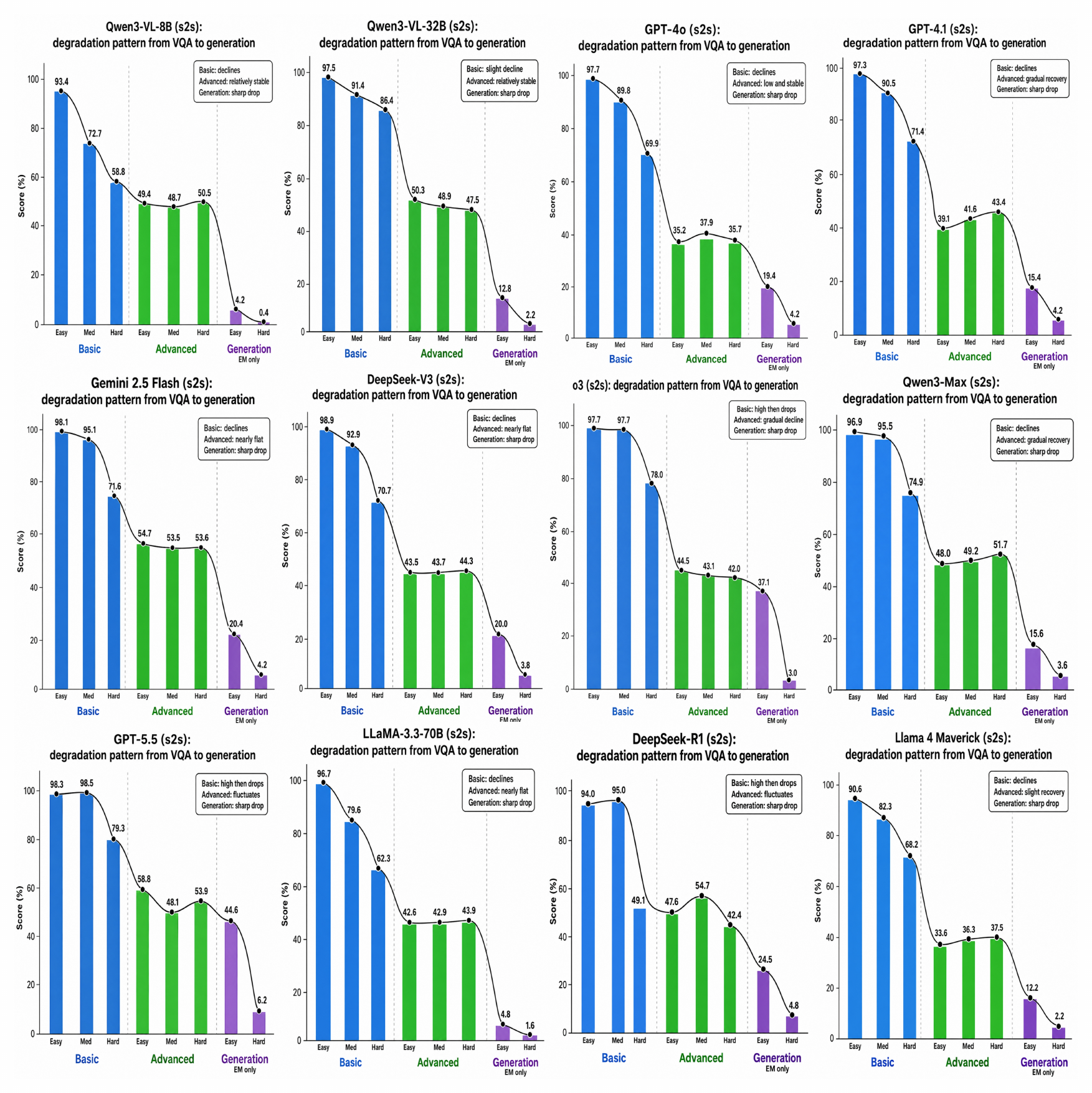}
  \caption{Cross-track degradation from VQA to Generation under 
  the s2s modality. For each model, blue bars show Basic VQA 
  accuracy, green bars show Advanced VQA accuracy, and purple 
  bars show Generation Exact Match (EM). The black curve connects 
  all scores to highlight the diagnostic performance cliff: models 
  that perform well on VQA, especially Basic Easy/Medium splits, 
  collapse when required to generate the edited molecule directly. 
  This pattern indicates that multiple-choice success often 
  reflects shortcut exploitation rather than robust atomic-level 
  R-group grounding.}
  \label{fig:s2s_degradation}
\end{figure*}

Figure~\ref{fig:s2s_degradation} visualizes the cross-track 
degradation pattern across representative s2s models, 
complementing the quantitative analysis.

\section{NOTA Robustness}
\label{app:nota}

\begin{table}[htb]
  \centering
  \small
  \setlength{\tabcolsep}{3.5pt}
  \begin{tabular}{@{}l cc c@{}}
    \toprule
    & \multicolumn{2}{c}{\textbf{NOTA is correct}} & \\
    \cmidrule(lr){2-3}
    \textbf{Model}
    & Struct.\ err & Non-exist R
    & \textbf{NOTA as distr.} \\
    \midrule
    \multicolumn{4}{l}{\textit{VLMs (img-smi)}} \\
    \addlinespace[2pt]
    Qwen3-VL-8B      & 61.0             & 75.8                            & 73.3 \\
    Qwen3-VL-32B     & 88.6             & 97.0                            & 72.1 \\
    Claude Sonnet 4.6  & 92.9             & \underline{98.0}                & 80.2 \\
    GPT-4o           & 90.0             & 94.9                            & 72.1 \\
    GPT-4.1          & 76.2             & 80.8                            & \textcolor{red}{\textbf{98.8}} \\
    GPT-5.5          & \textcolor{red}{\textbf{98.1}} & \textcolor{red}{\textbf{100.0}} & 76.7 \\
    Gemini 2.5 Flash & \underline{93.8} & \textcolor{red}{\textbf{100.0}} & 67.4 \\
    Llama 4 Maverick & 40.4             & 50.5                            & \underline{95.3} \\
    \rowcolor[gray]{0.93}
    \emph{Average}   & 80.1             & 86.6                            & 79.5 \\
    \midrule
    \multicolumn{4}{l}{\textit{LLMs (smi-smi)}} \\
    \addlinespace[2pt]
    DeepSeek-V4-Pro  & 80.6             & 88.5                            & 63.6 \\
    DeepSeek-R1      & \underline{98.4} & \textcolor{red}{\textbf{99.0}} & 28.8 \\
    LLaMA-3.3-70B    & 54.3             & 84.8                            & \textcolor{red}{\textbf{93.0}} \\
    o3               & \textcolor{red}{\textbf{100.0}} & \underline{97.1} & 58.3 \\
    Qwen3-Max        & 92.3             & 95.9                            & \underline{90.5} \\
    \rowcolor[gray]{0.93}
    \emph{Average}   & 85.1             & 93.1                            & 66.8 \\
    \bottomrule
  \end{tabular}
  \caption{%
    NOTA robustness (\%, img-smi for VLMs; smi-smi for LLMs).
    \textcolor{red}{\textbf{Red bold}}: best; \underline{underline}: second best.
  }
  \label{tab:nota}
\end{table}

\section{Advanced Property Reasoning}
\label{app:property}

\begin{table}[htb]
  \centering
  \small
  \setlength{\tabcolsep}{4pt}
  \begin{tabular}{@{}l ccc@{}}
    \toprule
    \textbf{Model}
    & \textbf{Physchem} & \textbf{Bioactivity} & \textbf{Indication} \\
    \midrule
    \multicolumn{4}{l}{\textit{VLMs (img-smi, Hard)}} \\
    \addlinespace[2pt]
    Qwen3-VL-8B      & 85.5             & 26.3                            & 83.7 \\
    Qwen3-VL-32B     & \textcolor{red}{\textbf{89.3}} & 26.7    & \underline{88.4} \\
    Claude Sonnet 4.6  & 66.7             & 23.9                            & 82.6 \\
    GPT-4o           & 86.2             & \textcolor{red}{\textbf{29.5}} & 82.1 \\
    GPT-4.1          & 84.9             & 26.0                            & \textcolor{red}{\textbf{88.9}} \\
    GPT-5.5          & 78.6             & \underline{27.4}               & 86.8 \\
    Gemini 2.5 Flash & \underline{88.7} & \underline{27.4}               & 83.7 \\
    Llama 4 Maverick & 79.2             & 24.9                            & 80.5 \\
    \rowcolor[gray]{0.93}
    \emph{Average}   & 82.4             & 26.5                            & 84.6 \\
    \midrule
    \multicolumn{4}{l}{\textit{LLMs (smi-smi, Hard)}} \\
    \addlinespace[2pt]
    DeepSeek-V4-Pro  & 70.7             & 28.7                            & \textcolor{red}{\textbf{87.9}} \\
    DeepSeek-R1      & 51.2             & \textcolor{red}{\textbf{33.0}} & 51.7 \\
    LLaMA-3.3-70B    & \underline{78.6} & 24.9                            & 43.2 \\
    o3               & 58.1             & 24.3                            & \underline{63.0} \\
    Qwen3-Max        & \textcolor{red}{\textbf{83.0}} & \underline{27.0} & 62.6 \\
    \rowcolor[gray]{0.93}
    \emph{Average}   & 68.3             & 27.6                            & 61.7 \\
    \bottomrule
  \end{tabular}
  \caption{%
    Advanced property reasoning accuracy (\%, Hard split) by property type.
    \textcolor{red}{\textbf{Red bold}}: best; \underline{underline}: second best.
  }
  \label{tab:property}
\end{table}

\section{Evaluation Scope of Chemical-domain Models}
\label{app:chemical_models}

Chemical-domain models are developed for different molecular understanding
objectives, and therefore their applicable evaluation settings differ from
those of general-purpose LLMs and VLMs.

\paragraph{Chemical VLMs.}
ChemVLM-8B and MolVL-7B are evaluated on the VQA track to assess molecular
recognition and candidate selection.
Their training objectives primarily focus on molecular representation
alignment and structure--text understanding rather than explicit
instruction-grounded molecular editing.
Advanced VQA introduces additional requirements, including physicochemical
property estimation, structure--activity relationship (SAR) understanding,
and biological activity interpretation, which are beyond the primary
objectives of existing chemical VLM pretraining.
Therefore, we focus their evaluation on Basic VQA and discuss Advanced VQA
performance separately in Appendix.

\paragraph{Markush-specific model.}
MarkushGrapher-2 is specifically designed for Markush structure
reconstruction and molecular generation, rather than general-purpose
molecular question answering or instruction following.
We therefore evaluate it on the Generation track, which partially overlaps
with its original objective of converting Markush representations into
instantiated molecular structures.
The reported evaluation uses molecular images as input, consistent with its
original image-based Markush reconstruction setting.

However, our Generation track additionally requires instruction-grounded
R-group resolution, where models must interpret natural-language editing
instructions and perform the corresponding substituent substitution.
This capability is not explicitly targeted by MarkushGrapher-2.
VQA evaluation is not directly comparable because it provides a predefined
candidate set and measures candidate-level discrimination, whereas
Generation evaluates autonomous molecular instantiation without candidate
constraints.




\section{Quantitative Analysis of Shortcut Control}
\label{app:shortcut}

To quantify whether the difficulty splits successfully control
similarity-based shortcuts, we define a per-question
\emph{similarity gap}:
\begin{equation}
  \Delta_{\text{sim}} = \mathrm{Tanimoto}(c,\,s)
    - \tfrac{1}{3}\!\sum_{w}\mathrm{Tanimoto}(w,\,s),
  \label{eq:simgap}
\end{equation}
where $c$ is the correct option, $w$ ranges over the three distractors,
and $s$ is the Markush scaffold using ECFP4 fingerprints (2048-bit).
A positive $\Delta_{\text{sim}}$ indicates that the correct option has
a scaffold-similarity advantage, creating a potential shortcut,
whereas values close to zero indicate that this structural cue is removed.

By construction, Easy-Basic yields a mean
$\Delta_{\text{sim}}{=}{+}0.33$, while Medium-Basic and Hard-Basic both
yield approximately $0.00$, showing that similarity-based shortcuts are
largely restricted to Easy-Basic.
Table~\ref{tab:simgap} reports accuracy across $\Delta_{\text{sim}}$ bins
for 8 VLMs under the img-smi setting.
Easy-Basic shows a clear downward trend as similarity decreases
(99.3\% at $\Delta_{\text{sim}}\geq0.50$ to 94.3\% at
$[-0.05,\,0)$), indicating that scaffold similarity provides useful
shortcuts in this setting.
In contrast, Hard-Basic remains nearly unchanged across similarity bins,
suggesting that its difficulty mainly arises from instruction-grounded
structural reasoning rather than distractor similarity.

\paragraph{Bootstrap significance of difficulty gaps.}
To verify that the Easy-to-Hard accuracy drop is not attributable to
sampling variability, we apply bootstrap resampling
($n{=}10{,}000$ iterations, seed~42) to each split independently.
Mean accuracies with 95\% confidence intervals are:
Easy-Basic $98.7\%$ [$98.22$, $99.08$],
Medium-Basic $93.7\%$ [$92.76$, $94.62$],
and Hard-Basic $62.4\%$ [$59.52$, $65.11$].
The Easy-to-Hard gap $\Delta_\text{EH}=36.3$ percentage points
(95\% CI [$33.51$, $39.10$], $p<0.001$, one-tailed bootstrap test),
confirming that the performance cliff is statistically significant.
The Easy-to-Medium gap is $5.0$ percentage points
(95\% CI [$3.96$, $5.98$], $p<0.001$), and the Medium-to-Hard gap is
$31.4$ percentage points
(95\% CI [$28.40$, $34.27$], $p<0.001$).

\paragraph{Correlation between similarity gap and accuracy.}
To further examine whether scaffold similarity predicts model accuracy,
we compute the Spearman correlation $\rho$ between per-question
$\Delta_{\text{sim}}$ and mean accuracy across 8 VLMs, with bootstrap
95\% confidence intervals ($n{=}10{,}000$).
We focus on VLMs because this analysis targets visual candidate
recognition under image-based input settings.
The correlations are:
Easy-Basic $\rho={+}0.138$
($p<0.001$, 95\% CI [$0.06$, $0.21$]),
Medium-Basic $\rho={+}0.025$
($p=0.501$, 95\% CI [$-0.05$, $0.10$]),
and Hard-Basic $\rho={-}0.008$
($p=0.815$, 95\% CI [$-0.07$, $0.06$]).
The significant positive correlation on Easy-Basic indicates that
questions with stronger scaffold-similarity advantages are easier for
models.
The near-zero, non-significant correlations on Medium- and Hard-Basic
confirm that distractor similarity no longer predicts accuracy after
shortcut control.

\begin{table}[t]
\centering
\small
\setlength{\tabcolsep}{6pt}
\begin{tabular}{lccc}
\toprule
\textbf{Gap} $\Delta_{\text{sim}}$
  & \textbf{Easy} & \textbf{Medium} & \textbf{Hard} \\
\midrule
$< -0.05$                            & ---  & 91.9 & 60.3 \\
$[-0.05,\;\;0.00)$                   & 94.3 & 93.3 & 63.4 \\
$[\phantom{-}0.00,\;\;0.05)$        & 95.7 & 93.6 & 61.2 \\
$[\phantom{-}0.05,\;\;0.10)$        & 98.4 & 98.4 & 62.1 \\
\midrule
$[\phantom{-}0.10,\;\;0.20)$        & 98.0 & ---  & ---  \\
$[\phantom{-}0.20,\;\;0.50)$        & 98.6 & ---  & ---  \\
$\geq 0.50$                         & 99.3 & ---  & ---  \\
\bottomrule
\end{tabular}
\caption{Accuracy (\%) by similarity gap $\Delta_{\text{sim}}$
(Eq.~\ref{eq:simgap}), averaged over 8 VLMs under the img-smi setting.
NOTA questions are excluded.
The lower rows correspond to Easy-Basic samples with larger
scaffold-similarity advantages, which are absent in Medium- and Hard-Basic
due to controlled distractor construction.
``---'': $n<10$.}
\label{tab:simgap}
\end{table}


\section{Evaluation Prompts}
\label{app:prompts}

All models are evaluated under a standardized zero-shot protocol with
temperature set to 0.
We describe the system prompt and user prompt templates for each evaluation
track below. The VQA track covers all four input-output modalities
(img-img, img-smi, smi-img, smi-smi); the Generation track covers
smi-smi and img-smi only.

\subsection{VQA Track}
\label{app:prompts:vqa}

\paragraph{System prompt (all VQA settings).}
\begin{quote}
\ttfamily\small
You are an expert chemist answering multiple choice questions.
YOUR ONLY OUTPUT MUST BE A SINGLE LETTER: A, B, C, or D.
DO NOT write any explanation, reasoning, analysis, or punctuation.
DO NOT start your response with words.
ONLY output one of these four characters: A B C D
\end{quote}

\paragraph{User prompt --- Basic, smi-smi / img-smi.}
The Markush structure is provided either as an image (img-smi) or as an
E-SMILES string (smi-smi); the four candidates are always provided as
SMILES strings.
\begin{quote}
\ttfamily\small
\{Markush image\} \textit{[img-smi only]}\\
Markush structure (SMILES): \{E-SMILES\} \textit{[smi-smi only]}\\[4pt]
Instruction: \{instruction\}\\[4pt]
\{question\}\\[4pt]
Candidate molecules (SMILES):\\
A: \{SMILES\_A\}\\
B: \{SMILES\_B\}\\
C: \{SMILES\_C\}\\
D: \{SMILES\_D\}\\[4pt]
\textit{[NOTA only]} Note: Option D is `None of the above'.
Select D if no option is chemically valid or correct.\\[4pt]
Your answer (single letter only, no explanation):
\end{quote}

\paragraph{User prompt --- Basic, smi-img / img-img.}
The four candidates are provided as molecule images appended after the text.
\begin{quote}
\ttfamily\small
\{Markush image\} \textit{[img-img only]}\\
Markush structure (SMILES): \{E-SMILES\} \textit{[smi-img only]}\\[4pt]
Instruction: \{instruction\}\\[4pt]
\{question\}\\[4pt]
The four candidate molecules are shown below as images (A, B, C, D).\\
\textit{[NOTA only]} Note: Option D is `None of the above'.
Select D if no option is chemically valid or correct.\\[4pt]
\{image\_A\} \{image\_B\} \{image\_C\} \{image\_D\}\\[4pt]
Your answer (single letter only, no explanation):
\end{quote}
For any candidate whose molecule is ``None of the above'' under an
image-candidate setting, the corresponding image slot is replaced by the
text line \texttt{Option \{label\}: None of the above}.

\paragraph{User prompt --- Advanced.}
Advanced tasks omit the \texttt{Instruction} line and replace it with a
property-based \{question\}. Indication/Mechanism questions present the
four options as \emph{R-groups} (labelled ``Candidate R-groups''),
whereas Physicochemical and Bioactivity questions present full
\emph{candidate molecules} (labelled ``Candidate molecules''). As in the
Basic task, candidates are rendered as SMILES in the smi-smi / img-smi
settings and as appended images in the img-img / smi-img settings; the
Markush input line additionally notes ``with an R-group position''.
Advanced tasks contain no NOTA variant.
\begin{quote}
\ttfamily\small
\{Markush image\} \textit{[img-smi only]}\\
Markush structure (SMILES) with an R-group position: \{E-SMILES\}
\textit{[smi-smi only]}\\[4pt]
\{property question\}\\[4pt]
Candidate molecules (SMILES): \textit{[Physicochemical / Bioactivity]}\\
Candidate R-groups (SMILES): \textit{[Indication/Mechanism]}\\
A: \{SMILES\_A\}\\
B: \{SMILES\_B\}\\
C: \{SMILES\_C\}\\
D: \{SMILES\_D\}\\[4pt]
Your answer (single letter only, no explanation):
\end{quote}
In the img-img / smi-img settings, the four option lines above are
instead appended as images after the question, exactly as in the Basic
image-candidate prompt.

\subsection{Generation Track}
\label{app:prompts:gen}

The Generation track is evaluated under two input modalities only,
smi-smi (E-SMILES in, SMILES out) and img-smi (image in, SMILES out);
there are no candidate options and hence no NOTA variant. Both
modalities share the same system prompt and differ only in how the
Markush structure is presented.

\paragraph{System prompt.}
\begin{quote}
\ttfamily\small
You are an expert chemist.
Given a Markush structure and an R-group substitution instruction,
generate the SMILES string of the resulting molecule.
Return ONLY the SMILES string enclosed in \textless{}smiles\textgreater{}
and \textless{}/smiles\textgreater{} tags.
Do not include any explanation, reasoning, or additional text.
Example format: \textless{}smiles\textgreater{}CCO\textless{}/smiles\textgreater{}
\end{quote}

\paragraph{User prompt --- smi-smi (E-SMILES input).}
The Markush structure is provided as an E-SMILES string.\footnote{The
field is labelled ``(SMILES)'' in the prompt for brevity, but the value
passed is the full E-SMILES string with R-group position annotations.}
\begin{quote}
\ttfamily\small
Markush structure (SMILES):\\
\{E-SMILES\}\\[4pt]
Instruction: \{instruction\}\\[4pt]
Generate the SMILES of the resulting molecule after applying this
instruction.
Return ONLY the SMILES enclosed in
\textless{}smiles\textgreater{}\textless{}/smiles\textgreater{} tags.
\end{quote}

\paragraph{User prompt --- img-smi (image input).}
The Markush structure is provided as a rendered image, prepended before
the text content.
\begin{quote}
\ttfamily\small
\{Markush image\}\\[4pt]
The image above shows a Markush structure with R-group position(s).\\[4pt]
Instruction: \{instruction\}\\[4pt]
Generate the SMILES of the resulting molecule after applying this
instruction.
Return ONLY the SMILES enclosed in
\textless{}smiles\textgreater{}\textless{}/smiles\textgreater{} tags.
\end{quote}

\paragraph{Output parsing.}
We extract the predicted SMILES from the model response in two stages.
First, we search for the \texttt{<smiles>...</smiles>} tag pattern
(case-insensitive, spanning newlines) and, if found, take its enclosed
content. If no tag is present, we fall back to the last non-empty line
of the response, accepting it as a SMILES only if it contains no
whitespace and is shorter than 300 characters; otherwise the prediction
is treated as empty (and thus invalid). The extracted string is then
canonicalized via RDKit before computing Exact Match, Scaffold Match,
and Tanimoto Similarity.


\section{Limitations}
\label{sec:limitations}

\paragraph{\textbf{Patent-level complexity.}}
Although R-GroundBench is constructed from real patent and literature
Markush structures, the benchmark focuses on well-defined R-group grounding
instances with controlled instructions and structured evaluation settings.
Real patent documents may contain more complex R-group definitions involving
long-range textual dependencies, conditional clauses, and heterogeneous
document layouts.
Nevertheless, the current benchmark follows common Markush description
patterns in patent claims and already reveals substantial grounding failures
in existing models.
Extending evaluation toward full patent-level retrieval and reasoning remains
an important future direction.

\paragraph{\textbf{Multiple-choice and generation scope.}}
The VQA track uses controlled candidate selection to isolate R-group grounding
ability, although multiple-choice settings may still permit limited
structure-level shortcuts.
The Generation track evaluates end-to-end molecular instantiation through
free-form SMILES generation, but it currently focuses on instruction-guided
R-group substitutions rather than unconstrained molecular discovery.
Future benchmarks could incorporate broader document-level reasoning and
more open-ended molecular design scenarios.

\section{The Use of Large Language Models}

We used large language models (GPT-4o and Claude Sonnet 4.6) only as general-purpose writing assistants to refine the clarity and fluency of the manuscript, such as improving grammar, style, and consistency of academic tone. All scientific ideas, methodological designs (including the  \bench{} framework, algorithms, and experiments), data collection, and analysis were conceived and executed entirely by the authors. Any LLM-generated or suggested text was critically reviewed and edited to ensure accuracy and faithfulness to our original contributions.

\end{document}